\pdfoutput=1
\documentclass[11pt]{article}

\usepackage{EMNLP2022}

\usepackage{times}
\usepackage{latexsym}
\usepackage{amsmath, amsfonts,rotating,mathrsfs}
\usepackage{multirow}
\usepackage[T1]{fontenc}
\usepackage[utf8]{inputenc}

\usepackage{microtype}

\usepackage{inconsolata}
\usepackage{xcolor}
\usepackage{tabularx,paralist}
\usepackage{array,booktabs,ragged2e}
\newcolumntype{R}[1]{>{\RaggedLeft\arraybackslash}p{#1}}
\usepackage{enumitem}

\title{ADDMU: Detection of Far-Boundary Adversarial Examples with Data and Model Uncertainty Estimation}

\author{Fan Yin \\
   University of California, Los Angeles \\
  \texttt{fanyin20@cs.ucla.edu} \\\And
  Yao Li \\
  University of North Carolina, Chapel Hill \\
  \texttt{yaoli@email.unc.edu} \\\AND
   Cho-Jui Hsieh \\
  University of California, Los Angeles \\
  \texttt{chohsieh@cs.ucla.edu} \\\And
    Kai-Wei Chang \\
  University of California, Los Angeles \\
  \texttt{ kwchang@cs.ucla.edu} \\}

\begin{document}
\maketitle
\begin{abstract}
Adversarial Examples Detection (AED) is a crucial defense technique against adversarial attacks and has drawn increasing attention from the Natural Language Processing (NLP) community. Despite the surge of new AED methods, our studies show that existing methods heavily rely on a shortcut to achieve good performance. In other words, current search-based adversarial attacks in NLP stop once model predictions change, and thus most adversarial examples generated by those attacks are located near model decision boundaries. To surpass this shortcut and fairly evaluate AED methods, we propose to test AED methods with \textbf{F}ar \textbf{B}oundary (\textbf{FB}) adversarial examples. Existing methods show worse than random guess performance under this scenario. To overcome this limitation, we propose a new technique, \textbf{ADDMU}, \textbf{a}dversary \textbf{d}etection with \textbf{d}ata and \textbf{m}odel \textbf{u}ncertainty, which combines two types of uncertainty estimation for both regular and FB adversarial example detection. Our new method outperforms previous methods by 3.6 and 6.0 \emph{AUC} points under each scenario. Finally, our analysis shows that the two types of uncertainty provided by \textbf{ADDMU} can be leveraged to characterize adversarial
examples and identify the ones that contribute most to model's robustness in adversarial training. 
%data and identify useful data that helps model robustness most when augmented to the training set. 

% Finally, we demonstrate two use cases of adversarial detection: first, it helps to better understand adversarial examples; second, it helps to identify adversarial examples that don't keep the semantics.

% \yli{A shorter version: (Hope it would be helpful! :)) Adversarial Examples Detection (AED) is a critical defense technique against adversarial attacks, and has drawn increasing attention from the Natural Language Processing (NLP) community. 
% However, our studies show that existing AED methods heavily rely on a shortcut to achieve good performance. The shortcut is that most adversarial examples generated by current attack methods locate near model decision boundaries. When tested against high probability (HP) adversarial examples, the existing methods show worse than random performance. To overcome this limitation, we propose a new technique, \textbf{ADDMU}, \textbf{a}dversary \textbf{d}etection with \textbf{d}ata and \textbf{m}odel \textbf{u}ncertainty, which combines data and model uncertainty estimation for good performance on both regular and HP adversarial examples. Finally, we demonstrate two use cases of adversarial detection: first, it provides better understanding of adversarial examples; second, it helps to identify adversarial examples that do not keep the semantics.}
\end{abstract}

\section{Introduction}
% \yli{From Yao: I think that capitalize the first letters when defining acronym is better. }

Deep neural networks (DNN) have achieved remarkable performance in a wide variety of NLP tasks. However, it has been shown that DNNs can be vulnerable to adversarial examples \citep{Jia2017AdversarialEF, alzantot2018generating, Jin2020IsBR}, i.e., perturbed examples that flip model predictions but remain imperceptible to humans, and thus impose serious security concerns about NLP models. 

To improve the robustness of NLP models, different kinds of techniques to defend against adversarial examples have been proposed~\citep{li2021searching}. In this paper, we study AED, which aims to add a detection module to identify and reject malicious inputs based on certain characteristics. Different from adversarial training methods~\citep{Madry2018TowardsDL, jia2019certified} which require re-training of the model with additional data or regularization, AED operates in the test time and can be directly integrated with any existing model. 

% . Those methods need to change the training stage of the NLP models, often introducing large overhead, so may not be a good  choice for real applications. Instead, we study Adversarial Example Detection (AED) in this paper, which aims to add a detection module to identify and reject malicious inputs based on certain characteristics. It only operates in the test time and can be directly integrated with any existing model to perform defense without re-training. 

%AED is one of the techniques to defend against adversarial examples, which identifies and rejects malicious inputs based on some 
%by their
% \yli{based on some}
%characteristics. Unlike other defense techniques like adversarial training \citep{Madry2018TowardsDL} which require re-training of the model with additional dataor regularization, AED operates on test time only, and can be directly integrated with a victim model to perform defense without re-training.

% Compared to other defenses like adversarial training  and certified robustness \citep{jia2019certified}, \yli{which require re-training of} the model with additional data or regularization, 

% AED is far more computational efficient, less harmful to clean accuracy, and can be directly integrated with a victim model to perform defense without re-training.

%only recently, people in NLP domain begin to realize the importance of AED \citep{zhou-etal-2019-learning, mozes-etal-2021-frequency, yoo-etal-2022-detection, Xie2022IdentifyingAA}, and propose several approaches
Despite being well explored in the vision domain \citep{Feinman2017DetectingAS, Raghuram2021AGF}, AED started to get attention in the field of NLP only recently. Many works have been proposed to conduct detection based on certain statistics~\citep{zhou-etal-2019-learning, mozes-etal-2021-frequency, yoo-etal-2022-detection, Xie2022IdentifyingAA}. Specifically, \citet{yoo-etal-2022-detection} propose a benchmark for AED methods and a competitive baseline by robust density estimation. However, by studying examples in the benchmark, we find that the success of some AED methods relies heavily on the shortcut left by adversarial attacks: 
%adversarial examples are mostly those that \emph{locate near model decision boundaries}, 
most adversarial examples \emph{are located near model decision boundaries}, i.e., they have small probability discrepancy between the predicted class and the second largest class. This is because when creating adversarial data, the searching process stops once model predictions changed. We illustrate this finding in Section \ref{2.2}.
%since the searching process in adversarial attacks stops once model predictions changed. 
%We empirically demonstrate this point in Section \ref{2.2} by showing that correctly detected adversarial examples are mainly those with small differences between the largest and the second largest softmax probability.

% We conduct quality checks to validate that FB adversarial examples do not suffer from obvious grammatical and semantic changes compared to regular ones.

%To avoid this shortcut and more accurately evaluate detection methods, we advocate for also evaluating on FB adversarial examples, 
To evaluate detection methods accurately, we propose to test AED methods on both regular adversarial examples and \textbf{F}ar-\textbf{B}oundary (\textbf{FB})\footnote{Other works may call this `High-Confidence'. We use the term `Far-Boundary' to avoid conflicts between `confidence' and the term `uncertainty' introduced later.} adversarial examples, which are created by continuing to search for better adversarial examples till a threshold of probability discrepancy is met.  Results show that existing AED methods perform worse than random guess on FB adversarial examples. %\citet{Athalye2018ObfuscatedGG} present similar observations in the vision domain. 
\citet{yoo-etal-2022-detection} recognize this limitation, but we find that this phenomenon is more severe than what is reported in their work. Thus, an AED method that works for FB attacks is in need.

% It is defined as the expected probability over the neighbors of the input on the original predicted class.
We propose ADDMU, an uncertainty estimation based AED method. The key intuition is based on the fact that adversarial examples lie off the manifold of training data and models are typically uncertain about their predictions of them. Thus, although the prediction probability is no longer a good uncertainty measurement when adversarial examples are far from the model decision boundary, there exist other statistical clues that give out the `uncertainty' in predictions to identify adversarial data. In this paper, we introduce two of them: \emph{data uncertainty} and \emph{model uncertainty}. Data uncertainty is defined as the uncertainty of model predictions over neighbors of the input.  Model uncertainty is defined as the prediction variance on the original input when applying Monte Carlo Dropout (MCD)~\citep{Gal2016DropoutAA} to the target model during inference time.
Previous work has shown that models trained with dropout regularization \citep{Srivastava2014DropoutAS} approximate the inference in Bayesian neural networks with MCD, where model uncertainty is easy to obtain \citep{Gal2016DropoutAA, Smith2018UnderstandingMO}. 
%Given the statistics for the two sources of uncertainty, 
Given the statistics of the two uncertainties, we apply p-value normalization~\citep{Raghuram2021AGF} and combine them with Fisher's method \citep{fisher1992statistical} to produce a stronger test statistic
for AED. To the best of our knowledge, we are the first work to estimate the uncertainty of Transformer-based models \citep{Shelmanov2021HowCI} for AED.

% The intuition is based on the observation that models are still \emph{uncertain} about their predictions for adversarial examples as they lie off the manifold of training data,
% %even they were assigned a high probability to the predicted class \citep{Smith2018UnderstandingMO}.
% even though they are assigned with high probabilities of the predicted class~\citep{Smith2018UnderstandingMO}.

%Model uncertainty refers to the softmax variance when applying Monte Carlo Dropout (MCD) \citep{Gal2016DropoutAA} to the target model during inference time. 

The advantages of our proposed AED method include: 1) it only operates on the output level of the model; 2) it requires little to no modifications to adapt to different architectures; 3) it provides an unified way to combine different types of uncertainties. Experimental results on four datasets, four attacks, and two models demonstrate that our method outperforms existing methods by 3.6 and 6.0 in terms of AUC scores on regular and FB cases, respectively. We also show that the two uncertainty statistics can be used to characterize adversarial data and select useful data for another defense technique, adversarial data augmentation (ADA). 

The code for this paper could be found at \url{https://github.com/uclanlp/AdvExDetection-ADDMU}

\section{A Diagnostic Study on AED Methods}
In this section, we first describe the formulation of adversarial examples and AED. Then, we show that current AED methods mainly act well on detecting adversarial examples near the decision boundary, but are confused by FB adversarial examples. 
%near the decision boundary adversarial examples.
\subsection{Formulation}
\noindent\textbf{Adversarial Examples.} Given an NLP model $f: \mathcal{X} \rightarrow \mathcal{Y}$, a textual input $x \in \mathcal{X}$, a predicted class from the candidate classes $y \in \mathcal{Y}$, and a set of boolean indicator functions of constraints, $\mathcal{C}_i: \mathcal{X} \times \mathcal{X} \rightarrow \{0, 1\}, i=1, 2,\cdots, n$. An (untargeted) adversarial example $x^* \in \mathcal{X}$ satisfies:
\begin{equation*}
        f\left(x^*\right) \neq f\left(x\right), \\
    \mathcal{C}_i\left(x, x^*\right) = 1, i = 1, 2, \cdots, n.
\end{equation*}
Constraints are typically grammatical or semantic similarities between original and adversarial data. For example, \citet{Jin2020IsBR} conduct part-of-speech checks and use Universal Sentence Encoder \citep{cer2018universal} to ensure semantic similarities between two sentences. 

\noindent\textbf{Adversarial Examples Detection (AED)}
The task of AED is to distinguish adversarial examples from natural ones, based on certain characteristics of adversarial data.  We assume access to 1) the victim model $f$, trained and tested on clean datasets $\mathcal{D}_{train}$ and $\mathcal{D}_{test}$; 2) an evaluation set $\mathcal{D}_{eval}$ ; 3) an auxiliary dataset $\mathcal{D}_{aux}$ contains only clean data. $\mathcal{D}_{eval}$ contains equal number of adversarial examples $\mathcal{D}_{eval-adv}$ and natural examples $\mathcal{D}_{eval-nat}$. $\mathcal{D}_{eval-nat}$ are randomly sampled from $\mathcal{D}_{test}$. $\mathcal{D}_{eval-adv}$ is generated by attacking a disjoint set of samples from $\mathcal{D}_{eval-nat}$ on $\mathcal{D}_{test}$. See Scenario 1 in \citet{yoo-etal-2022-detection} for details. We use a subset of $\mathcal{D}_{train}$ as $\mathcal{D}_{aux}$. We adopt an unsupervised setting, i.e., the AED method is not trained on any dataset that contains adversarial examples.

\subsection{Diagnose AED Methods}
\label{2.2}

We define examples \emph{near model decision boundaries} to be those whose output probabilities for the predicted class and the second largest class are close. 
%Regular iterative adversarial attacks stop once they find adversarial examples that change model predictions. Therefore, we suspect regular attacks are mostly generating adversarial data near the boundaries, and AED methods could rely on this property to make decisions. 
Regular iterative adversarial attacks stop once the predictions are changed. Therefore, we suspect that regular attacks are mostly generating adversarial examples near the boundaries, and existing AED methods could rely on this property to detect adversarial examples.

% \yli{From Yao: Figure 1 shows the results of two attacks: TextFooler and Pruthi and does not include \citep{yoo-etal-2022-detection}. Why we say Figure 1 verifies the method in \citep{yoo-etal-2022-detection}?}

\begin{figure}[t]
\centering
\includegraphics[scale=0.355]{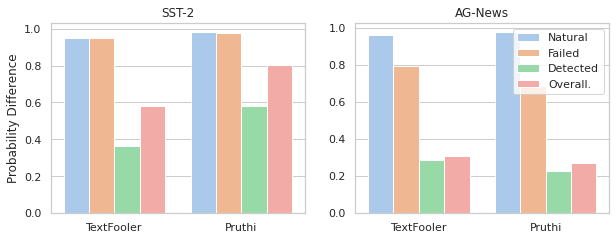}
\vspace{-10pt}
\caption{The probability difference between the predicted class and the second largest class on natural examples, adversarial examples that the detector failed, succeed, and in total. The X-axis is the attack. The Y-axis is the difference. Correctly detected adversarial examples have relatively small probability difference.}
\label{probdiff}
\end{figure}

Figure \ref{probdiff} verifies this for the state-of-the-art unsupervised AED method \citep{yoo-etal-2022-detection} in NLP, denoted as \textbf{RDE}. Similar trends are observed for another baseline. 
%The X-axis is attack method. We show two popular methods, TextFooler \citep{Jin2020IsBR} and Pruthi \citep{pruthi2019combating}. The Y-axis is the probability difference between the predicted class and the second largest class. We show the average probability difference on adversarial examples that the detector fails to identify (\textcolor{CornflowerBlue}{Failed}), successfully detected (\textcolor{Peach}{Detected}), and the overall average probability difference (\textcolor{LimeGreen}{Overall}). There is a clear trend that adversarial examples with smaller probability differences between the predicted class and the second largest class are more likely to be identified as natural examples, and correctly detected adversarial examples have high probability difference.
The X-axis shows two attack methods: TextFooler~\citep{Jin2020IsBR} and Pruthi~\citep{pruthi2019combating}. The Y-axis represents the probability difference between the predicted class and the second largest class. Average probability differences of natural examples (Natural), and three types of adversarial examples are shown: RDE fails to identify (Failed), successfully detected (Detected), and overall (Overall). There is a clear trend that successfully detected adversarial examples are those with small probability differences while the ones with high probability differences are often mis-classified as natural examples. This finding shows that these AED methods identify examples near the decision boundaries, instead of adversarial examples.

% There is a clear trend that adversarial examples with small probability differences are more likely to be detected, while the ones with high probability differences are often mis-classified as natural examples.

% (\textcolor{CornflowerBlue}{Failed}), successfully detected (\textcolor{Peach}{Detected}), and overall (\textcolor{LimeGreen}{Overall}). 

\begin{table}[t]
\renewcommand{\arraystretch}{0.85}
\begin{center}
\small
\begin{tabular}{m{1.6cm}<{\centering}m{1.0cm}<{\centering}m{1.0cm}<{\centering}m{1.0cm}<{\centering}m{1.0cm}<{\centering}}
% & \multicolumn{2}{c}{\bf{PTB-SD3.5}}
\toprule 
& \multicolumn{2}{c}{\bf{RDE}} & \multicolumn{2}{c}{\bf{DIST}}\\
\cmidrule{2-5}
\textbf{Data-attack} & \textbf{Regular} & \textbf{FB} & \textbf{Regular} & \textbf{FB}  \\ 

\toprule 
SST2-TF & 72.8/86.5 & 45.0/81.5 & 73.4/87.9 & 26.3/81.6 \\
SST2-Pruthi & 55.1/80.6 & 30.8/72.6 & 61.4/85.3 & 26.5/74.6 \\
\midrule
Yelp-TF & 79.2/89.6 & 44.6/82.7 & 80.3/90.6 & 64.3/86.2 \\
Yelp-Pruthi & 64.8/88.0 & 47.9/85.2 & 72.2/89.2 & 55.2/84.9 \\
\bottomrule\hline
\end{tabular}
\vspace{-10pt}
\end{center}
\caption{F1/AUC scores of two SOTA detection methods on \textbf{Regular} and \textbf{FB} adversarial examples. RDE and DIST perform worse than random guess (F1=50.0) on FB adversarial examples.}
\label{FBnearrandom}
\end{table}

To better evaluate AED methods,
%we propose to avoid the above shortcut by evaluating detection methods with FB adversarial examples, 
we propose to avoid the above shortcut by testing detection methods with FB adversarial examples, which are generated by continuously searching for adversarial examples until a prediction probability threshold is reached. We simply add another goal function to the adversarial example definition to achieve this while keep other conditions unchanged:
\begin{multline*}
    f\left(x^*\right) \neq f\left(x\right),  p\left(y = f\left(x^*\right) \mid x^* \right) \geq \epsilon \\
    \mathcal{C}_i\left(x, x^* \right) = 1, i = 1, 2, \cdots, n.
\end{multline*}
% \begin{gather}
%         f\left(x^*\right) \neq f\left(x\right),  p\left(y = f\left(x^*\right) \mid x^* \right) \geq \epsilon \\
%     \mathcal{C}_i\left(x, x^* \right) = 1, i = 1, 2, \cdots, n.
% \end{gather}
$p\left(y = f\left(x^*\right) \mid x^* \right)$ denotes the predicted probability for the adversarial example. $\epsilon$ is a manually defined threshold. We illustrate the choice of $\epsilon$ in Section \ref{ExpSetup}. In Table \ref{FBnearrandom}, it shows that the existing competitive methods (RDE and DIST) get lower than random guess F1 scores when evaluated with FB adversarial examples.

% State-of-the-art AED methods in NLP either adopt robust, class conditional density estimation on the feature space \citep{yoo-etal-2022-detection}, or use the discrepancy between class conditional distances between the evaluated point to training points (a baseline proposed in this paper) to do detection.

% They estimate parameters of a class conditional multivariate Gaussian with features of examples in $\mathcal{D}_{train}$. Then, adversarial examples would have smaller Mahalanobis scores using the estimated Gaussian distribution \citep{Lee2018ASU}.

\subsection{Quality Check for FB Attacks}
\begin{table}[t]
\renewcommand{\arraystretch}{0.85}
\begin{center}
\small
\begin{tabular}{m{1.7cm}<{\centering}m{0.9cm}<{\centering}m{0.9cm}<{\centering}m{0.9cm}<{\centering}m{0.9cm}<{\centering}}
% & \multicolumn{2}{c}{\bf{PTB-SD3.5}}
\toprule 
& \multicolumn{2}{c}{\bf{Grammar}} & \multicolumn{2}{c}{\bf{Semantics}}\\
\cmidrule{2-5}
\textbf{Data} & \textbf{Regular} & \textbf{FB} & \textbf{Regular} & \textbf{FB}  \\ 

\toprule 
SST-2 & 1.117 & 1.129 & 3.960 & 3.900 \\
Yelp & 1.209 & 1.233 & 4.113 & 4.082 \\

\bottomrule\hline
\end{tabular}
\end{center}
\vspace{-10pt}
\caption{Quality checks for FB adversarial examples. The results on each dataset are averaged over examples from three attacks: TextFooler, BAE, Pruthi, and their FB versions. The numbers for Grammar columns are the relative increases of errors of perturbed examples w.r.t. original examples. The numbers for Semantics columns are the averaged rates that the adversarial examples preserve the original meaning evaluated by humans. The quality of adversarial examples do not degrade much with the FB version of attacks.}
\label{FBcheck}
\end{table}

% \begin{table*}[t]
% \renewcommand{\arraystretch}{0.85}
% \begin{center}
% \small
% \begin{tabular}{m{1.25cm}m{7.8cm}}
% \toprule
% \multicolumn{2}{c}{BAE attack on SST-2, ADDMU fails to detect}\\
% \midrule
% \multicolumn{2}{c}{Groundtruth Label changed : Positive $\rightarrow$ Negative} \\
% \midrule
% Original & Seattle may have just won the 2014 Super Bowl, but the Steelers still [[rock]] with six rings, Baby!!! Just stating what all Steeler fans know: a Steel Dynasty is still unmatched no matter what team claims the title of current Super Bowl Champs.. Go Steelers!!!
% Regular & Seattle may have just won the 2014 Super Bowl, but the Steelers still [[trembles]] with six rings, Baby!!! Just stating what all Steeler fans know: a Steel Dynasty is still unmatched no matter what team claims the title of current Super Bowl Champs.. Go Steelers!!! \\
% FB & Seattle may have just won the 2014 Super Bowl, but the Steelers [[again]] [[trembles]] with six rings, Baby!!! Just stating what all Steeler fans know: a Steel Dynasty is still unmatched no matter what team claims the title of current Super Bowl Champs.. Go Steelers!!! \\
% \bottomrule
% \end{tabular}
% \end{center}
% \vspace{-6pt}
% \caption{Why detector performance varies among attacks? This might because attacks already flip groundtruth labels of the examples. We show the detector performance (F1) and the proportion of adversarial examples that have their sentiments changed according to humans on correctly and wrongly detected sets. }
% \label{qualcheck}
% \end{table*}

We show that empirically, the quality of adversarial examples do not significantly degrade even searching for more steps and stronger FB adversarial examples. We follow \citet{Morris2020ReevaluatingAE} to evaluate the quality of FB adversarial examples in terms of grammatical and semantic changes, and compare them with regular adversarial examples. We use a triple $\left(x, x_{adv}, x_{FB-adv}\right)$ to denote the original example, its corresponding regular adversarial and FB adversarial examples. For grammatical changes, we conduct an automatic evaluation with LanguageTool \citep{naber2003rule} to count grammatical errors and report the relative increase of errors of perturbed examples w.r.t. original examples. For semantic changes, we do a human evaluation using Amazon MTurk \footnote{We pay workers 0.05 dollars per HIT. Each HIT takes approximately 15 seconds to finish. So, we pay each worker 12 dollars per hour. Each HIT is assigned three workers.}. We ask the workers to rate to what extent the changes to $x$ preserve the meaning of the sentence, with scale 1 (`Strongly disagree') to 5 (`Strongly agree'). Results are summarized in Table \ref{FBcheck}. The values are averaged over three adversarial attacks, 50 examples for each. We find that the FB attacks have minimal impact on the quality of the adversarial examples. We show some examples on Table \ref{Appendix: qualcheck}, which qualitatively demonstrate that it is hard for humans to identify FB adversarial examples.

\section{Adversary Detection with Data and
Model Uncertainty (ADDMU)}
Given the poor performance of previous methods on FB attacks, we aim to build a detector that can handle not only regular but also FB adversarial examples. We propose ADDMU, an uncertainty estimation based AED method by combing two types
%sources 
of uncertainty: model uncertainty and data uncertainty. We expect the adversarial examples to have large values for both. The motivation of using uncertainty is that 
%models can still be uncertain about their predictions even they assign a high value to predicted class. 
models can still be uncertain about their predictions even when they assign a high probability of predicted class to an example.
% Note that although we use the same names for the two uncertainties as in \citep{Xiao2019QuantifyingUI}, the definitions are different. \yli{From Yao: I think we don't need to mention this.}
We describe the definitions  and estimations of the two uncertainties, and how to combine them.

\subsection{Model Uncertainty Estimation}
Model uncertainty represents the uncertainty when predicting a single data point with randomized models. \citet{Gal2016DropoutAA} show that model uncertainty can be extracted from DNNs trained with dropout and inference with MCD without any modifications of the network. This is because the training objective with dropout minimizes the Kullback-Leibler divergence between the posterior distribution of a Bayesian network and an approximation distribution. We follow this approach and define the model uncertainty as the softmax variance when applying MCD during test time.

Specifically, given a trained model $f$, we do $N_m$ stochastic forward passes for each data point $x$. The dropout masks of hidden representations for each forward pass are i.i.d sampled from a Bernolli distribution, i.e., $z_{lk} \sim \emph{Bernolli}\left(p_m\right)$ where $p_m$ is a fixed dropout rate for all layers, $z_{lk}$ is the mask for neuron $k$ on layer $l$. Then, we can do a Monte Carlo estimation on the softmax variance among the $N_m$ stochastic softmax outputs. 
%Denote the probability of predicting the i-th class on j-th forward pass as $p_{ij}$ and the mean probability for the i-th class over $T$ passes as $\hat{p}_i = \sum_{j=1}^T p_{ij}$, the model uncertainty (MU) is:
% (\yli{From Yao: since it's mean, the definition is missing the fraction $\frac{1}{T}$. Also, since we're using $N_m$ in previous definition and the following equation, why not use it here as well? $T$ is not defined but $N_m$ is defined. Since it's a mean stat, bar might be better than hat. Proposal: and the mean probability for the i-th class over $N_m$ passes as $\bar{p}_i = \sum_{j=1}^{N_m} p_{ij}$,} )
Denote the probability of predicting the input as the $i$-th class in the $j$-th forward pass as $p_{ij}$ and the mean probability for the $i$-th class over $N_m$ passes as $\bar{p}_i = \frac{1}{N_m}\sum_{j=1}^{N_m} p_{ij}$,  the model uncertainty (MU) can be computed by
%\begin{equation}
%    MU\left(x\right) = \frac{1}{|\mathcal{Y}|}\sum_{j=1}^{|\mathcal{Y}|}\frac{1}{N_m}\sum_{i=1}^{N_m}\left(p_{ij} - \hat{p}_i\right)^2.
%\end{equation}
\begin{equation*}
    MU\left(x\right) = \frac{1}{|\mathcal{Y}|}\sum\nolimits_{i=1}^{|\mathcal{Y}|}\frac{1}{N_m}\sum\nolimits_{j=1}^{N_m}\left(p_{ij} - \bar{p}_i\right)^2.
\end{equation*}

\subsection{Data Uncertainty Estimation}
% Model uncertainty represents the uncertainty when predicting a single data point with randomized models. \yli{From Yao: suggest to move this sentence to the beginning of section 3.1.}
Data uncertainty quantifies the predictive probability distribution of a fixed model over the neighborhood of an input point.

Specifically, similar to the model uncertainty estimation, we do $N_d$ stochastic forward passes. But instead of randomly zeroing out neurons in the model, we fix the trained model and construct a stochastic input for each forward pass by masking out input tokens, i.e., replacing each token in the original input by a special token 
%following a Bernolli distribution $\emph{Bernolli}\left(p_d\right)$, where $p_d$ is the masking probability. 
with probability $p_d$.
%We use the expected ( 1- maximum softmax probability) over the $N_d$ forward passes as the quantification for data uncertainty. 
The data uncertainty is estimated by the mean of (1 $-$ maximum softmax probability) over the $N_d$ forward passes.
Denote the $N_d$  stochastic inputs as $x_1, x_2, \cdots, x_{N_d}$, the original prediction as $y$, and the predictive probability of the original predicted class as $p_y\left( \cdot \right)$, the Monte Carlo estimation on data uncertainty (DU) is:
\begin{equation*}
    DU\left(x\right) = \frac{1}{N_d}\sum\nolimits_{i=1}^{N_d} \left(1 - p_y\left( x_i \right)\right).
\end{equation*}

% Maximum softmax probability is a type of model confidence estimation, but might not be reliable enough as shown in previous work. 

%  Specifically,  The neighborhood is constructed by replacing each work in the sentence by a special token with some probabilities.

% The intuition is that 

\subsection{Aggregate Uncertainties with Fisher's Method}

We intend to aggregate the two uncertainties described above to better reveal the low confidence of model's prediction on adversarial examples.  We first normalize the uncertainty statistics so that they follow the same distribution. Motivated by \citet{Raghuram2021AGF} where the authors normalize test statistics across layers 
%using p-value, 
by converting them to p-values,
we also adopt the 
%p-value 
same method
to normalize the two uncertainties. By definition, a p-value computes the probability of a test statistic being at least as extreme as the target value. 
%It will transform 
The transformation will convert
any test statistics into a uniformly distributed probability. We construct empirical distributions for MU and DU by calculating the corresponding uncertainties for each example on the auxiliary dataset $\mathcal{D}_{aux}$, denoted as $T_{mu}$, and $T_{du}$. Following the null hypothesis \emph{$H_0$: the data being evaluated comes from the clean distribution}, we can calculate the p-values based on model uncertainty ($q_m$) and data uncertainty ($q_d$) by:
\begin{equation*}
    \begin{split}
        q_m\left(x\right) &= \mathbb{P}\left(T_{mu} \geq MU\left(x\right) \mid H_0 \right),\\
        q_d\left(x\right) &= \mathbb{P}\left(T_{du} \geq DU\left(x\right) \mid H_0 \right).
    \end{split}
\end{equation*}
The smaller the values $q_m$ and $q_d$, the higher the probability of the example being adversarial.
% \begin{align}
%     q_u\left(x\right) = \mathbb{P}\left(T_{mu} \geq MU\left(x\right) \mid H_0 \right)\\
%     q_m\left(x\right) = \mathbb{P}\left(T_{du} \geq DU\left(x\right) \mid H_0 \right).
% \end{align}

%Then, given $q_m$ and $q_d$, we combine them into a single value using the Fisher's method for combined probability test \citep{fisher1992statistical}. 
Given $q_m$ and $q_d$, we combine them into a single p-value using the Fisher's method to do combined probability test~\citep{fisher1992statistical}. Fisher's method indicates that under the null hypothesis, the sum of the log of the two p-values follows a $\chi^2$ distribution with 4 degrees of freedom. We use $q_{agg}$ to denote the aggregated p-value. Adversarial examples should have smaller $q_{agg}$, where
%\begin{equation}
$    \text{log} q_{agg} = \text{log} q_m + \text{log} q_d.$
%\end{equation}
% Although Fisher's method assumes independence across the tests, which might not hold in this case, we find that it works well empirically. \yli{(From Yao: maybe we can remove this sentence.)}

\section{Experiments}
We first describe the experimental setup (Section \ref{ExpSetup}), then present our results on both regular and FB AED (Section \ref{ExpRes}). Results show that our ADDMU outperforms existing methods by a large margin under both scenarios.
\subsection{Experimental Setup}
\label{ExpSetup}
\noindent{\bf Datasets and victim models.} 
%We experiment with classification tasks on different domains include sentiment analysis SST-2 \citep{socher2013recursive}, Yelp \citep{zhang2015character}, topic classification AGNews \citep{zhang2015character}, and natural language inference SNLI \citep{Bowman2015ALA}. 
We conduct experiments on classification tasks in different domains, including sentiment analysis SST-2 \citep{socher2013recursive}, Yelp \citep{zhang2015character}, topic classification AGNews \citep{zhang2015character}, and natural language inference SNLI \citep{Bowman2015ALA}.
We generate both regular and FB adversarial examples on the test data of each dataset with two word-level attacks: TextFooler (TF) \citep{Jin2020IsBR}, BAE \citep{garg2020bae}, and two character-level attacks: Pruthi \citep{pruthi2019combating}, and TextBugger (TB) \citep{Li2019TextBuggerGA}. We only consider the examples that are predicted correctly before attacks. The numbers of evaluated examples vary among 400 to 4000 across datasets. See Appendix \ref{Appendix:ExpSetupDetails}. For FB adversarial examples, we choose the $\epsilon$ so that adversarial examples have approximately equal averaged prediction probability with natural data. Specifically, $\epsilon = 0.9$ for SST-2, Yelp, AGNews, and $\epsilon = 0.7$ for SNLI. We mainly experiment with two Transformer-based victim models, BERT \citep{devlin2019bert} and RoBERTa \citep{Liu2019RoBERTaAR} as they are widely adopted in the current NLP pipelines and show superior performance than other architectures. More details are presented in Appendix \ref{Appendix:ExpSetupDetails}. In Appendix \ref{Appendix:LSTM}, we also present some simple experiments with BiLSTM. 

\noindent{\bf Baselines.} We compare \textbf{ADDMU} with several unsupervised AED methods. 1) \textbf{MSP:} \citet{Hendrycks2017ABF} use the Maximum Softmax Probability (MSP) for detection; 2) \textbf{PPL:} GPT-2 large \citep{Radford2019LanguageMA} as a language model to measure the perplexity of the input; 3) \textbf{FGWS:} \citet{mozes-etal-2021-frequency} measure the difference in prediction probability after replacing infrequent words of the inputs with frequent words and find that adversarial examples have higher performance change; 4) \textbf{RDE:} \citet{yoo-etal-2022-detection} fit class conditional density estimation with Kernel PCA~ \citep{Schlkopf1998NonlinearCA} and Minimum Covariance Determinant~\citep{Rousseeuw1984LeastMO} in the feature space and use the density scores; 5) \textbf{DIST:} we propose a distance-based baseline that uses the difference between class conditional, averaged K nearest distances. See Appendix \ref{Appendix:dist} for details.

Unsupervised AED methods assign a score to each evaluated data. Then, a threshold is selected based on the maximum False Positive Rate (FPR) allowed, i.e., the rate of mis-classified natural data.

% Both \textbf{RDE} and \textbf{DIST} can be enhanced by calculating the average score over the neighborhood of the input as in data uncertainty. We call those methods \textbf{RDE-aug} and \textbf{DIST-aug}.

\noindent{\bf Implementation Details.}
For \textbf{FGWS} and \textbf{RDE}, we follow the hyper-parameters in their papers to reproduce the numbers. For \textbf{DIST} and \textbf{ADDMU}, we attack the validation set and use those examples to tune the hyper-parameters. See Appendix \ref{Appendix:implement} for details. Specifically, for \textbf{DIST}, we use 600 neighbors. For \textbf{ADDMU}, we find $N_m=10$, $p_m = 0.2$ for MU works well for all datasets. For DU, we find that it is beneficial to ensemble different mask rates for text classification tasks, we set $N_d = 100$ in total, and 25 for each $p_d \in \{0.1, 0.2, 0.3, 0.4\}$ for all the text classification tasks, $N_d = 25$, $p_d = 0.1$ for SNLI.

\noindent{\bf Metrics.} In the main experiments, we select the threshold at maximum FPR=0.1. A lower FPR represents a more practical case where only a small proportion of natural samples are mis-classified as adversarial samples. Following the setup in \citet{Xu2018FeatureSD} and \citet{yoo-etal-2022-detection}, we report True Positive Rate (TPR), i.e., the fraction of the real adversarial examples out of predicted adversarial examples, and F1 score at FPR=0.1, and Area Under the ROC curve (AUC), which measures the area under the TPR and FPR curve. For all the metrics, the higher the better.

\begin{table*}[!t]
\renewcommand{\arraystretch}{0.8}
\centering
\small
\begin{tabular}{m{1.58cm}|m{1.38cm}|m{.55cm}<{\centering}m{.55cm}<{\centering}m{.65cm}<{\centering}|m{.55cm}<{\centering}m{.55cm}<{\centering}m{.65cm}<{\centering}|m{.55cm}<{\centering}m{.55cm}<{\centering}m{.65cm}<{\centering}|m{.55cm}<{\centering}m{.55cm}<{\centering}m{.65cm}<{\centering}}
  \toprule
  &  & \multicolumn{3}{c}{\bf{SST-2}} & \multicolumn{3}{c}{\bf{AGNews}} & \multicolumn{3}{c}{\bf{Yelp}} &
  \multicolumn{3}{c}{\bf{SNLI}}
\\
\cmidrule{1-6}\cmidrule{7-10}\cmidrule{11-14}
   \bf{Attacks} & \bf{Methods} & \bf{TPR}  & \bf{F1} & \bf{AUC} & \bf{TPR}  & \bf{F1} & \bf{AUC} & \bf{TPR}  & \bf{F1} & \bf{AUC}  & \bf{TPR}  & \bf{F1} & \bf{AUC}\\
  \toprule
  \multirow{8}{4em}{{TF}} &

 PPL&31.2& 44.2& 72.4& 76.1& 81.8& 91.1& 45.7& 58.8& 79.3& 40.2& 53.6& 78.0 \cr 
 & FGWS & 62.9 & 72.8 & 76.5 & 83.0& 86.3& 85.5 & 67.1& 72.7& 80.6 & 48.5 & 55.4 & 72.2 \cr 
  &MSP&64.0& 73.6& 88.0& 95.2& 92.8& 97.5& 73.9& 80.4& 90.6& 56.7& 68.0& 83.6 \cr 
 & RDE&62.9& 72.8& 86.5& 96.0& 93.2& 97.0& 72.0& 79.2& 89.6& 46.3& 59.3& 81.0 \cr 
 &DIST&64.0& 73.4& 87.9& 94.5& 92.4& 95.9& 73.8& 80.3& 90.6& 37.2& 50.4& 74.5 \cr 
 &ADDMU& \bf 67.1& \bf 75.8&\bf  88.8& \bf 99.2& \bf 94.9& \bf 98.6& \bf 78.7&\bf  83.5& \bf 91.6& \bf 68.9& \bf 77.0& \bf 89.7 \cr
 \midrule
   \multirow{8}{4em}{{TF-FB}} &

 PPL&41.9& 55.2& 80.6& 83.3& 86.3& 93.9& 49.9& 62.4& 81.6& 44.1& 57.2& 79.2 \cr 
 &FGWS & 61.8& 72.0 & 77.9& 84.8& 87.1& 88.1& 72.2& 78.0 & 89.4 & 52.1& 59.6 & 78.4 \cr 
  &MSP&31.2& 44.2& 81.9& 82.0& 85.4& 91.5& 66.0& 75.0& 87.1& 26.8& 39.2& 75.1 \cr 
 &RDE&31.9& 45.0& 81.5& 71.9& 79.1& 92.5& 31.5& 44.6& 82.7& 43.1& 56.4& 79.6 \cr 
 &DIST&20.7& 26.3& 81.6& 66.6& 75.4& 91.8& 54.8& 64.3& 86.2& 27.2& 39.6& 69.9 \cr 
 &ADDMU&\bf 62.0& \bf 72.2& \bf 88.0 &\bf  97.5& \bf 94.0& \bf 97.8& \bf 72.8& \bf 79.7& \bf 89.7& \bf 53.6& \bf 65.8& \bf 87.5 \cr 
  \midrule
  \midrule
   \multirow{8}{4em}{{BAE}} &
 PPL&19.7& 30.4& 66.2& 30.9& 44.0& 71.8& 23.6& 35.3& 70.1& 24.8& 36.8& 68.1 \cr 
 &FGWS & 37.6&  51.0& 64.2& 64.7& 74.2& 72.5& 54.9& 66.7&68.0&31.2&44.0& 67.9 \cr 
  &MSP&45.1& 58.3& 79.0& 96.0& 93.4& 96.0& 68.3& 76.7& 89.5& 41.4& 54.7& 71.4 \cr 
 &RDE&44.2& 57.3& 79.3& 96.4& \bf 93.7& 96.3& 65.2& 74.5& 89.1& 41.7& 55.0& 76.8 \cr 
 &DIST&44.9& 57.3& 78.9& 94.2& 91.9& 96.2& 68.0& 76.2& 89.4& 36.8& 49.7& 67.9 \cr 
 &ADDMU&\bf 45.9& \bf 58.9&\bf  82.3&\bf  96.4&  93.5& \bf 97.3& \bf 72.5& \bf 79.5& \bf 90.1& \bf 48.2& \bf 61.0& \bf 81.0 \cr 
  \midrule
   \multirow{8}{4em}{{BAE-FB}} &
 PPL&26.0& 38.2& 70.5& 45.5& 58.7& 79.6& 28.5& 41.3& 73.0& 24.9& 37.0& 67.9 \cr 
 &FGWS & 20.4 & 31.4& 57.1& 72.6& 79.6& 78.2& 51.9& 64.3& 65.9 &32.9& 47.5& 63.4 \cr 
  &MSP&12.8& 21.1& 70.4& 79.2& 83.8& 91.2& 69.1& 77.2& 88.3& 18.3& 28.6& 62.6 \cr 
 &RDE&19.5& 30.2& 72.5& 68.8& 77.0& 91.2& 66.4& 75.4& 88.1& 34.6& 47.9& 74.0 \cr 
 &DIST&17.7& 26.1& 70.1& 64.9& 68.1& 91.4& 69.7& 77.3& 88.4& 29.5& 42.3& 62.9 \cr 
 &ADDMU&\bf 51.4&\bf  64.1&\bf  84.6&\bf  83.7&\bf  85.9& \bf 94.1&\bf  76.3& \bf 81.9&\bf  90.6&\bf  34.9& \bf 48.4&\bf  76.0 \cr 
  \midrule
  \midrule
   \multirow{8}{4em}{{Pruthi}} &
 PPL&29.7& 42.9& 71.9& 31.0& 44.0& 70.7& 35.3& 48.7& 72.9& 54.9& 66.6& 85.5 \cr 
  &MSP&53.2& 65.2& 82.6& 75.7& 81.9& 91.5& 65.4& 74.7& 88.7& 22.5& 33.9& 69.2 \cr 
 &RDE&41.4& 55.1& 80.6& 77.4& 82.8& 92.4& 52.6& 64.8& 88.0& 34.6& 47.8& 76.5 \cr 
 &DIST&55.0& 61.4& 82.9& 77.8& 82.0& 92.1& 66.7& 72.2& 88.2& 23.6& 35.2& 65.1 \cr 
 &ADDMU&\bf 55.9& \bf 67.4& \bf 85.4&\bf  96.7&\bf  93.9&\bf  97.4&\bf  78.8& \bf 83.7& \bf 91.8& \bf 55.7& \bf 67.1& \bf 86.0 \cr 
   \midrule
   \multirow{8}{5em}{{Pruthi-FB}} &
 PPL&28.6& 41.6& 72.3& 27.8& 40.4& 71.6& 37.3& 50.8& 73.3& 37.2& 50.6& 76.3 \cr 
  &MSP&31.1& 44.4& 73.8& 49.4& 62.2& 84.5& 51.5& 63.9& 85.4& 10.2& 17.0& 64.5 \cr 
 &RDE&20.0& 30.8& 72.6& 59.5& 70.4& 87.6& 34.3& 47.9& 85.2& 31.2& 44.2& 74.9 \cr 
 &DIST&23.3& 26.5& 74.6& 55.1& 61.6& 87.2& 54.5& 55.2& 84.9& 21.6& 32.8& 63.3 \cr 
 &ADDMU&\bf 56.2& \bf 68.7& \bf 85.8& \bf 80.4& \bf 84.9& \bf 95.0&\bf  68.7& \bf 77.0& \bf 90.7& \bf 44.9& \bf 58.0& \bf 82.5 \cr 
  \midrule
  \midrule
   \multirow{8}{4em}{{TB}} &
 PPL&30.8& 43.7& 76.1& 74.0& 80.5& 90.3& 56.9& 68.2& 84.4& 56.0& 67.5& 84.3 \cr 
  &MSP&72.3& 79.0& 90.5& 95.6& 93.0& 97.3& 70.4& 78.1& 89.8& 66.4& 75.1& 89.0 \cr 
 &RDE&72.4& 79.6& 89.6& 96.1& 93.3& 96.9& 66.2& 75.2& 89.2& 51.8& 64.1& 83.0 \cr 
& DIST&72.4& 78.6& 90.6& 95.6& 92.8& 96.2& 70.2& 77.9& 90.2& 50.7& 62.7& 82.6 \cr 
 &ADDMU&\bf 73.3&\bf  80.0&\bf  90.9&\bf  99.0& \bf 94.8& \bf 98.4& \bf 70.8& \bf 78.3& \bf 91.0&\bf  69.0& \bf 77.1& \bf 90.6 \cr 
    \midrule
   \multirow{8}{4em}{{TB-FB}} &
 PPL&36.0& 49.4& 80.2& 82.9& 86.0& 94.2& 60.6& 71.1& 85.8& 48.9& 61.6& 76.3 \cr 
  &MSP&34.8& 48.2& 83.0& 81.1& 84.9& 91.2& 70.0& 77.8& 88.4& 34.7& 48.0& 81.5 \cr 
 &RDE&29.5& 42.5& 82.1& 68.9& 77.1& 91.7& 63.9& 73.5& 88.4& 47.8& 60.6& 82.2 \cr 
 &DIST&34.3& 44.0& 82.6& 63.4& 72.9& 91.5& 69.8& 77.6& 89.3& 40.8& 53.9& 79.0 \cr 
 &ADDMU&\bf 50.5& \bf 62.9& \bf 86.1& \bf 94.2& \bf 92.6& \bf 96.9& \bf 74.8& \bf 81.0& \bf 90.8& \bf 51.1& \bf 63.6& \bf 87.0 \cr 
  
  \bottomrule\hline

\end{tabular}
\caption{Detection performance of regular and FB adversarial examples (*-FB) against BERT on SST-2, AGNews, Yelp, and SNLI. Our proposed ADDMU outperforms other methods by a large margin, especially on FB adversarial examples. We occlude FGWS under character-level attacks, Pruthi and TextBugger, as it is designed for word-level detection. The best performance is bolded. Results are averaged over three runs with different random seeds.}
\label{MainRes}
\end{table*}
\subsection{Results}
\label{ExpRes}
Performances of AED methods on BERT are presented in Table \ref{MainRes}. We average the results among three runs with different random seeds. See Appendix \ref{Appendix:Roberta} for the results on RoBERTa.

\noindent\textbf{Detector performance.} Our proposed ADDMU achieves the best performance on both regular and FB adversarial examples under the three metrics (TPR, F1, AUC) on the four datasets, which demonstrates the effectiveness of ADDMU. Further, ADDMU preserves more than 90\% of the performance or even achieves better results, e.g SST-2-Pruthi and Yelp-BAE, under FB adversarial attacks, which shows that ADDMU is not affected by FB attacks.

The performances of MSP, DIST, and RDE are severely degraded under FB attacks. This demonstrates that those methods can be fooled and circumvented by carefully designed attacks. Under regular attacks, the performances of RDE and DIST are worse than the baseline MSP in most cases, which simply uses the maximum softmax probability for detection. One explanation is that those class conditional methods are just approximating softmax probabilities so might not be as effective as MSP in detecting near the decision boundary examples.

%Finally, PPL and FGWS are also not affected much by FB adversarial examples. But, their performance are usually not sufficient for detection, and FGWS is only applicable to word-level attacks.
Finally, PPL and FGWS are also not severely affected by FB attacks. However, FGWS is only applicable to word-level attacks. Also, PPL and FGWS are not effective enough in general.

\noindent\textbf{Ablation study.}
%The value for data and model uncertainty (\textbf{DU} and \textbf{MU}) before aggregation can also be used as features and detect adversarial examples. And, both \textbf{RDE} and \textbf{DIST} can be enhanced to tackle FB adversarial examples by calculating the average score over the neighborhood of the input using the same random masking in data uncertainty.
Data uncertainty (\textbf{DU}) and model uncertainty (\textbf{MU}) can also be used as features in detection separately. Also, both \textbf{RDE} and \textbf{DIST} can be enhanced by calculating the average score over the neighborhood of the input using the same random masking technique as used in data uncertainty estimation.
We denote them as \textbf{RDE-aug} and \textbf{DIST-aug}. 
%In this part, we demonstrate the effectiveness of uncertainty aggregation and the augmenting technique by comparing ADDMU with DU, MU, RDE-aug, and DIST-aug. Full results are shown in Appendix \ref{appendix:ablation}. We show a representative proportion of the results in Table \ref{Ablation} and summarize the findings here. 
In this part, we study the effectiveness of uncertainty aggregation and neighbor augmentation by comparing ADDMU with DU and MU, and by comparing RDE and DIST with RDE-aug and DIST-aug. Full results are shown in Appendix \ref{appendix:ablation}. We show a representative proportion of the results in Table \ref{Ablation}. The summary of findings are discussed in the following.

%We find that the aggregation of two uncertainties, i.e., ADDMU, achieves the best values in 70 out of the 96 metric scores. DU and MU are the best in 12 scores each. 
We find that ADDMU, the aggregation of two uncertainties, achieves the best results in 70 out of the 96 metric scores.
DU and MU are the best in 12 scores each.
This shows that the combination of the two uncertainties provides more information to identify adversarial examples. 
%We also observe that RDE-aug and DIST-aug are in general better than RDE and DIST, especially in FB cases, which demonstrates the effectiveness of considering neighbors of the input for detection. Finally, we also notice that on SNLI, DU values are typically less useful, and thus the combination of DU and MU performs slightly worse than MU. We suspect this is because the NLI task requires more sophisticated neighborhood construction method to generate meaningful neighbors for data uncertainty estimation. We leave this to future work. 
We also observe that on SNLI, DU values are typically less useful, and thus the combination of DU and MU performs slightly worse than MU. One explanation is that the SNLI task requires more sophisticated neighborhood construction method to generate meaningful neighbors in data uncertainty estimation. Finally, we also notice that RDE-aug and DIST-aug are in general better than RDE and DIST, especially under FB attacks, which demonstrates the effectiveness of neighbor augmentation.

\begin{table}[t]
\renewcommand{\arraystretch}{0.85}
\begin{center}
\small
\begin{tabular}{m{0.1cm}<{\centering}|m{1.25cm}<{\centering}|m{0.5cm}<{\centering}m{0.5cm}<{\centering}m{0.5cm}<{\centering}m{0.5cm}<{\centering}m{0.5cm}<{\centering}m{0.5cm}<{\centering}}
% & \multicolumn{2}{c}{\bf{PTB-SD3.5}}
\toprule 
&  & \multicolumn{3}{c}{\textbf{AGNews}} & \multicolumn{3}{c}{\textbf{SNLI}}\\
\cmidrule{1-8}
& Method & TPR & F1& AUC & TPR & F1 & AUC\\
\cmidrule{1-8}
\multirow{7}{4em}{\rotatebox[origin=c]{90}{TF}} & RDE& 96.0& 93.2& 97.0& 46.3& 59.3& 81.0 \cr 
 &RDE-aug& 97.4& 94.0& 97.4& 41.0& 54.3& 79.9 \cr 
 &DIST& 94.5& 92.4& 95.9& 37.2& 50.4& 74.5 \cr 
 &DIST-aug& 94.0& 92.0& 96.9& 38.3& 51.5& 75.2 \cr 
 &MU& 82.0& 85.4& 94.5& 65.1& 74.4& 89.1 \cr 
 &DU& 98.9& 94.6& 98.3& 59.6& 70.3& 85.6 \cr 
 &ADDMU&\bf 99.2&\bf 94.9&\bf 98.6&\bf 68.9&\bf 77.0& \bf 89.7 \cr
\bottomrule\hline
\end{tabular}
\end{center}
\vspace{-6pt}
\caption{
%Ablation study on ADDMU and comparisons with enhanced baselines.
Ablation study on effect of uncertainty aggregation and neighbors augmentation against TextFooler. 
}
\label{Ablation}
\end{table}

\noindent\textbf{Why do detection results vary among datasets and attacks?} Among different attacks, we find that Pruthi is the hardest to detect, followed by BAE. 
%However, there is no clear trend whether word-level or character-level attacks are harder to detect.
However, there is no obvious difference between detection performances against word-level and character-level attacks.
Also, attacks on the sentence pair task (SNLI) are in general harder to detect. 
%Thus, future work could focus more on the more challenging task of detecting adversarial examples on sentence pair tasks, like SNLI.
Thus, future work could focus more on improving the performance of detecting adversarial examples in sentence pair tasks, like SNLI.

We investigate why the detection performances vary among attacks. Our hypothesis is that attacks on some datasets fail to be imperceptible and have changed the groundtruth label for an input. 
Thus, these `adversarial' (can not be called adversarial any more as they do not meet the definition of being imperceptible) examples actually lie close to the training manifold of the target class. Therefore, AED methods find it hard to detect those examples. 
To verify this assumption, we choose two tasks (SST-2 and Yelp) and two attacks (TF and BAE) to do sentiment analysis.
We ask Amazon MTurk workers \footnote{Also 0.05 dollar per HIT, but each HIT takes around 10 seconds to finish. Each HIT is assigned three workers.} to re-label \emph{positive} or \emph{negative} for attacked examples.
Then, we summarize the proportion of examples that workers assign opposite groundtruth labels in correctly and wrongly detected groups. As shown in Table \ref{whyvary}, there is an obvious correlation between bad performance and the number of `adversarial' examples whose groundtruth labels changed. For example, ADDMU performs weak on detecting BAE attacks on SST-2 (58.9 F1), but it turns out that this is because more than half of the examples already have their groundtruth labels flipped.
%We show a case in example in Table \ref{whyvary}. This shows that adversarial attack algorithms need to be improved to generate more natural adversarial examples, which would also benefit the develop of AED methods.
We give one example in Table \ref{whyvary}. This shows that adversarial attacks need to be improved to retain the semantic meaning of the original input.

\begin{table}[t]
\renewcommand{\arraystretch}{0.85}
\begin{center}
\small
\begin{tabular}{m{1.45cm}<{\centering}|m{0.9cm}<{\centering}|m{1.4cm}<{\centering}m{1.4cm}<{\centering}}
% & \multicolumn{2}{c}{\bf{PTB-SD3.5}}
\toprule 
& F1 & Correct & Wrong \\
\toprule 
SST-2 TF & 75.8 & 0.129 & 0.360 \\
SST-2 BAE  & 58.9 & 0.136 & 0.597 \\
Yelp TF & 83.5 & 0.211 & 0.411 \\
Yelp BAE & 79.5 & 0.229 & 0.425 \\
\bottomrule\hline
\end{tabular}
\begin{tabular}{m{1.25cm}m{4.8cm}}
\toprule
\multicolumn{2}{c}{BAE attack on SST-2, ADDMU fails to detect}\\
\midrule
\multicolumn{2}{c}{Groundtruth Label changed : Positive $\rightarrow$ Negative} \\
\midrule
Original & Most new movies have a \textcolor{red}{bright} sheen. \\
Attacked & Most new movies have a \textcolor{red}{bad} sheen. \\
\bottomrule
\end{tabular}
\end{center}
\vspace{-6pt}
\caption{Why detector performance varies among attacks? This might because attacks already flip groundtruth labels of the examples. We show the detector performance (F1) and the proportion of adversarial examples that have their sentiments changed according to humans on correctly and wrongly detected sets. }
\label{whyvary}
\end{table}

\section{Characterize Adversarial Examples}
In this section, we explore how to characterize adversarial examples by the two uncertainties.

\begin{figure}[t]
\centering
\includegraphics[scale=0.16]{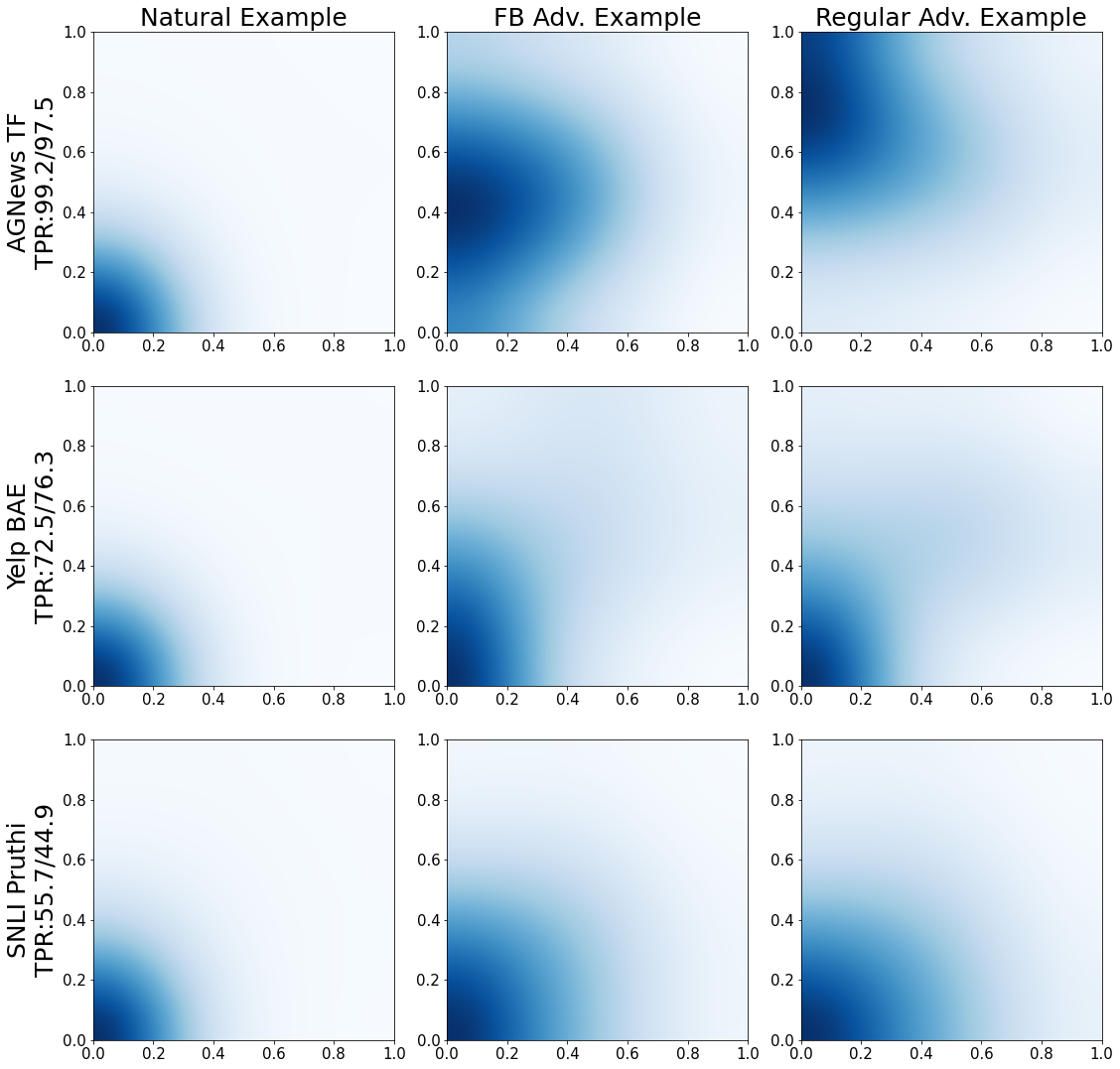}
\vspace{-6pt}
\caption{
%MU-DU heatmaps for three attacks with natural, FB adversarial, and regular adversarial examples. The X-axis is the MU value. The Y-axis is the DU value. Attack types and ADDMU performance (TPR: Regular Adv./FB Adv.) are shown on the left column.
MU-DU heatmaps based on natural and regular/FB adversarial examples generated from three attacks. X-axis: MU value; Y-axis: DU value. Attack types and ADDMU performance are labeled on the left. TPR: Regular Adv./FB Adv.
}
\label{DU-MUmap}
\end{figure}
%Regarding different MU and DU values as X-axis and Y-axis, we visualize input examples in a MU-DU data map. As in Figure \ref{DU-MUmap}, we construct heatmaps for natural, FB adversarial, and regular adversarial examples on three attacks, AGNews TF, Yelp BAE, and SNLI Pruthi. 
\noindent\textbf{MU-DU Data Map}
Plotting a heatmap with MU on X-axis and DU on Y-axis, we visualize data in terms of the two uncertainties. We show in Figure~\ref{DU-MUmap} the heatmaps with natural data, FB and regular adversarial examples generated from three attacks on three datasets (AGNews TF, Yelp BAE, SNLI Pruthi). The performance of ADDMU varies on the three attacks, as shown on the left of Figure~\ref{DU-MUmap}. 

We find that natural examples center on the bottom left corner of the map, representing low MU and DU values. This phenomenon does not vary across datasets. Whereas for FB and regular adversarial examples, they have larger values on at least one of the two uncertainties. 
%In the best case (AGNews TF, the first row), the center of adversarial examples in the MU-DU map moves upward and rightward. BAE examples on Yelp have larger DU values in general, as reflected by the deeper blue shadow on the top half of the maps on the second row. 
When ADDMU performs best (AGNews TF, the first row), the center of adversarial examples in the MU-DU map is relatively rightward and upward compared to other cases.
For maps on the third row, the shadow stretches along the MU axis, indicating that Pruthi examples on SNLI have relatively large MU values.

\noindent\textbf{Identifying Informative ADA Data}
ADA is another adversarial defense technique, which augments the training set with adversarial data and re-train the victim model to improve its robustness. In this part, we show that our ADDMU provides information to select adversarial data that is more beneficial to model robustness. We test it with TF on SST-2. The procedure is as follows: since SST-2 only has public training and validation sets, we split the original training set into training (80\%) and validation set (20\%), and use the original validation set as test set. We first train a model on %clean training data. 
the new training set.
Then, we attack the model on validation data and compute DU and MU values for each adversarial sample. We sort the adversarial examples according to their DU and MU values and split them by half into four disjoint sets: \textbf{HDHM} (high DU, high MU), \textbf{HDLM} (high DU, low MU), \textbf{LDHM} (low DU, high MU), and \textbf{LDLM} (low DU, low MU). We augment the clean training set with each of these sets and retrain the model. As a baseline, we also 
%compare with augmenting with the whole validation adversarial examples (All).
test the performance of augmenting with all the adversarial examples generated from the validation set (All). We report clean accuracy (\textbf{Clean \%}), the number of augmented data (\textbf{\#Aug}), attack success rate (\textbf{ASR}), and the average query number (\textbf{\#Query}) for each model.

The results are in Table \ref{ADATest}. We find that the most 
%informative 
helpful
adversarial examples are with \emph{low DU} and \emph{high MU}. Using those samples, we achieve better ASR and clean accuracy than augmenting with 
%the whole validation adversarial examples with only 25\% of the total examples in the whole set.
the whole validation set of adversarial examples, with only one quarter of the amount of data.
%The second informative ones are those with high DU and high MU. It is intuitive to see that examples with low DU and low MU are less useful as they are more similar to clean data.
It is expected that examples with low DU and low MU are less helpful as they are more similar to the clean data.
%We have similar observations on the FB version of TF. We also discuss the augmentation for regular and FB attacks. See Appendix XX
Similar observations are found in the FB version of TF attacks. We also compare augmentations with regular and FB adversarial examples. See details in Appendix \ref{Appendix:ADA}.

\begin{table}[t]
\renewcommand{\arraystretch}{0.85}
\begin{center}
\small
\begin{tabular}{R{1.6cm} p{1.20cm} p{0.70cm} p{1.20cm} p{0.80cm}}
% & \multicolumn{2}{c}{\bf{PTB-SD3.5}}
\toprule 
SST-2 TF &  \textbf{Clean \%} & \textbf{\#Aug}& \textbf{ASR} & \textbf{\#Query} \\
\toprule 
\multicolumn{1}{l}{BERT} &  92.8 & 0 & 94.31\% & 98.51 \\
+ All   & 92.4 & 11199 & 87.36\% & 66.31 \\
+ LDLM  & 91.7 & 2800& 90.62\% & 108.06 \\
+ HDLM  & 92.4 & 2800& 88.59\% & 111.26 \\
+ LDHM  & \textbf{92.9} & 2799& \textbf{85.05\%} & 115.30 \\
+ HDHM  & 91.9 & 2799 & 87.07\% & \textbf{119.92} \\
\bottomrule\hline
\end{tabular}
\vspace{-6pt}
\end{center}
\caption{ADA performances of different types of augmented data. We find that adversarial examples with low DU and high MU are most useful for ADA.}
\label{ADATest}
\end{table}

\section{Related Work}
\noindent{\bf Adversarial Detection.} Adversarial examples detection has been well-studied in the image domain \citep{Feinman2017DetectingAS, Lee2018ASU, Ma2018CharacterizingAS, Xu2018FeatureSD, Roth2019TheOA, Li2021DetectingAE, Raghuram2021AGF}. Our work aligns with 
\citet{Feinman2017DetectingAS, Li2021DetectingAE, Roth2019TheOA} that introduce uncertainty estimation or perturbations as features to detect adversarial examples. We postpone the details to Appendix \ref{Appendix:relatedwork}, but focus more on the AED in NLP domain.
 
In the NLP domain, there are less work exploring AED. \citet{zhou-etal-2019-learning} propose DISP that learns a BERT-based discriminator to defend against adversarial examples. \citet{mozes-etal-2021-frequency} propose a word-level detector FGWS that leverages the model confidence drop when replacing infrequent words in the input with frequent ones and surpass DISP. \citet{pruthi2019combating} combat character-level attacks with word-recognition models. More recently, \citet{yoo-etal-2022-detection} propose a robust density estimation baseline and a benchmark for evaluating AED methods. There are other works like~\citet{Xie2022IdentifyingAA, biju-etal-2022-input, Wang2022DetectingTA, Mosca2022ThatIA},  that leverage other features or train a detector. We show limitations of these works on FB adversarial examples and propose our ADDMU that overcomes this limitation.

\noindent{\bf Other Defenses against Attacks.} AED is a category of approaches to defending against adversarial attacks. Other methods are also considered. \citet{Jin2020IsBR, yin2020on, Si2021BetterRB} do \emph{ADA} that augments original training datasets with adversarial data for better robustness. \citet{Madry18Towards, Miyato2017AdversarialTM, Zhu2020FreeLBEA, Zhou2020DefenseAA} conduct \emph{adversarial training} which is formulated as a min-max problem. Recently, several works perform certified robustness defense with either interval bound propagation  \citep{Huang2019AchievingVR, jia2019certified, Shi2020Robust}, or randomized smoothness \citep{Ye2020SAFERAS}. In this work, we connect our AED method with ADA by selecting more informative data to augment.
\section{Conclusion}
We proposed ADDMU, an uncertainty-based approach for both regular and FB AED. We began by showing that existing methods are significantly affected by FB attacks. Then, we show that ADDMU is minimally impacted by FB attacks and outperforms existing methods by a large margin. We further showed ADDMU characterizes adversarial data and provides information on how to select useful augmented data for improving robustness.

\section*{Acknowledgement}
We thank anonymous reviewers, UCLA PLUS-
Lab and UCLA-NLP for their helpful feedback.
This work is partially supported by DMS-2152289, DMS-2134107, IIS-2008173,  IIS-2048280, Cisco Faculty Award, and a Sloan Research Fellow.

\section*{Limitations}
We summarize the limitations of this paper in this section.
\begin{enumerate}
    \item We only test the AED methods under classification tasks. This is because we find that the attacks on other tasks like language generation are not well-defined, for example what would be the goal function of attacks on a language generation task? Is minimizing the BLEU score sufficient? It is hard to conduct detection when there is no standard for a valid adversarial example. Future work might come up with attacks for diverse tasks first and propose corresponding AED methods.
    \item More experiments should be conducted to analyze the FB adversarial examples, including its characteristics and the security concerns it imposes to DNNs. However, given the time and space limitations, we are not able to do that.
    \item Our method has slightly more hyperparameters to tune (four in total), and requires a bit more time to finish one detection. But, we confirm that it is in an acceptable range.
\end{enumerate}

\bibliography{custom, newref, nlp, ref}
\bibliographystyle{acl_natbib}

\clearpage
\appendix
\begin{table*}[t] 
\begin{center}
\small
\begin{tabular}{m{1.1cm}|m{12.0cm}|m{0.8cm}}
\toprule
Attacks & Examples & Prob.\cr
\toprule
Original & Seattle may have just won the 2014 Super Bowl, but the Steelers still [[rock]] with six rings, Baby!!! Just stating what all Steeler fans know: a Steel Dynasty is still unmatched no matter what team claims the title of current Super Bowl Champs.. Go Steelers!!! & 100\% \cr
\midrule
Regular & Seattle may have just won the 2014 Super Bowl, but the Steelers still [[trembles]] with six rings, Baby!!! Just stating what all Steeler fans know: a Steel Dynasty is still unmatched no matter what team claims the title of current Super Bowl Champs.. Go Steelers!!! & 57\%\cr
\midrule
FB & Seattle may have just won the 2014 Super Bowl, but the Steelers [[again]] [[trembles]] with six rings, Baby!!! Just stating what all Steeler fans know: a Steel Dynasty is still unmatched no matter what team claims the title of current Super Bowl Champs.. Go Steelers!!! & 95\%\cr
\midrule
\midrule
Original & Cisco invests \$12 million in Japan R amp;D center On Thursday, the [[company]] announced it will invest \$12 million over the next five years in a new research and development center in Tokyo. & 71\% \cr
\midrule
Regular & Cisco invests \$12 million in Japan R amp;D center On Thursday, the [[firm]] announced it will invest \$12 million over the next five years in a new research and development center in Tokyo. & 63\%\cr
\midrule
FB & Cisco invests \$12 million in Japan R amp;D center On Thursday, the company [[mentioned]] it will invest \$12 million over the next five [[decades]] in a new research and development [[centre]] in Tokyo. & 95\%\cr
\midrule
\midrule
Original & King Pong Draws Fans Spike TV's Video [[Game]] Awards Show attracts big-name celebrities and bands but gives the fans the votes. & 93\% \cr
\midrule
Regular & King Pong Draws Fans Spike TV's [[tv]] Game Awards Show attracts big-name celebrities and bands but gives the fans the votes. & 57\%\cr
\midrule
FB & King Pong Draws Fans Spike TV's Video [[play]] Awards Show attracts big-name celebrities and bands but gives the fans the votes. & 90\%\cr

\bottomrule
\end{tabular}

\end{center}
\caption{Examples of the changes made by regular and far-boundary adversarial examples. The last column shows the prediction probability on the predicted class. We can see that it would still be hard for humans to identify the changes made by far boundary examples. It is necessary to propose an automatic way to detect far boundary adversarial examples.}
\label{Appendix: qualcheck}
\end{table*}
\section{Regular vs. FB Adversarial Examples}
In this section, we qualitatively shows some cases of far-boundary adversarial examples in Table \ref{Appendix: qualcheck}. We show that it is hard for human beings to identify such far-boundary examples, which calls for an automatic way to do the detection.

\section{Experimental Setup Details}
\label{Appendix:ExpSetupDetails}
\subsection{Datasets and Target Models}
We conduct experiments on four datasets, SST-2, Yelp-Polarity, AGNews, and SNLI. Statistics about those datasets are summarized on Table \ref{DataStatistics}. All those datasets are available at Huggingface Datasets \citep{lhoest-etal-2021-datasets}. Our target models are BERT \citep{devlin2019bert} and RoBERTa \citep{Liu2019RoBERTaAR}. We use the public accessible BERT-base-uncased and RoBERTa-base models fine-tuned on the above datasets provided by TextAttack \citep{morris2020textattack} to benefit reproducibility. The performance of those models are summarized on Table \ref{Appendix:ModelPerform}.
\begin{table}[h]
\centering
\begin{tabular}{m{1.0cm}m{2.8cm}<{\centering}m{1.15cm}<{\centering}m{.95cm}<{\centering}}
  \toprule
  Dataset & Train/Dev/Test & Avg Len & \#Labels \cr
  \toprule
  SST-2 & 67.3k/0.8k/- & 19.2 & 2\cr
  Yelp & 560k/-/38k & 152.8 & 2\cr
  AGNews & 120k/-/7.6k & 35.5 & 4\cr
  SNLI & 550k/10k/20k & 8.3 & 3\cr

  \bottomrule\hline
\end{tabular}
\caption{Data Statistics of the four datasets.}
\label{DataStatistics}
\end{table}

\begin{table}[h]
\centering
\begin{tabular}{m{1.3cm}m{1.0cm}<{\centering}m{1.15cm}<{\centering}m{1.2cm}<{\centering}m{1.0cm}<{\centering}}
  \toprule
  Dataset & SST-2 & Yelp & AGNews & SNLI\cr
  \toprule
  BERT & 92.43 & 96.30 & 94.20 & 89.40 \cr
  RoBERTa & 94.04 &  - & 94.70 & - \cr

  \bottomrule\hline
\end{tabular}
\caption{BERT-base-uncased and RoBERTa-base accuracy on the four datasets. TextAttack does not have public model for RoBERTa fine-tuned on Yelp and SNLI.}
\label{Appendix:ModelPerform}
\end{table}

\subsection{Attacks and Statistics}
We consider four attacks. TextFooler \citep{Jin2020IsBR}, BAE \citep{garg2020bae}, Pruthi \citep{pruthi2019combating}, and TextBugger \citep{Li2019TextBuggerGA}. TextFooler and BAE are word-level attacks. Pruthi and TextBugger are character-level attacks. For BAE, we use BAE-R, i.e., replace a word with a substitution. For attacks on SNLI, we only perturb the hypothesis sentence. For FB attacks, as stated in the main paper, we add another goal function to make sure the softmax probability of the attacked class is larger than a threshold $\epsilon$. We select $\epsilon = 0.9$ for SST-2, Yelp, AGNews, and $\epsilon = 0.7$ for SNLI. We implement those attacks with TextAttack, with the default hyperparameter settings. Please refer to the document of TextAttack for details. Here we report the after-attack accuracy (\textbf{Adv. Acc}), the attack success rate (\textbf{ASR}), the number of queries (\textbf{\#Query}), and the number of adversarial examples we select (\textbf{\#Adv}) for each attack on each dataset, as well as for FB attacks. Notice, the total evaluted examples will be twice the number of adversarial examples. See Table \ref{appendix:attacks} and Table \ref{appendix:attackshp}.

\begin{table}[t]
\renewcommand{\arraystretch}{0.85}
\begin{center}
\small
\begin{tabular}{m{1.3cm}<{\centering}|m{1.3cm}<{\centering}m{0.9cm}<{\centering}m{1.0cm}<{\centering}m{1.1cm}<{\centering}}
\toprule
\toprule 
\bf TextFooler & Adv. Acc &ASR\% & \#Query & \#Adv\\
\toprule 
SST-2 & 4.5 & 95.1 & 95.3 & 1290\\
Yelp & 6.0 & 93.8 & 475.7 & 738\\
AGNews & 17.7 & 81.4 & 333.5 & 1625\\
SNLI & 3.0 & 96.7 & 58.5 & 2222\\
\bottomrule\hline
\end{tabular}
\begin{tabular}{m{1.3cm}<{\centering}|m{1.3cm}<{\centering}m{0.9cm}<{\centering}m{1.0cm}<{\centering}m{1.1cm}<{\centering}}
\toprule 
\bf BAE & Adv. Acc &ASR\% & \#Query & \#Adv\\
\toprule 
SST-2 & 38.3 & 58.9 & 60.8 & 412 \\
Yelp & 44.9 & 53.7 & 319.9 & 1039\\
AGNews & 81.5 & 14.3 & 122.5 & 278\\
SNLI & 32.5 & 64.0 & 43.4 & 1605\\
\bottomrule\hline
\end{tabular}
\begin{tabular}{m{1.3cm}<{\centering}|m{1.3cm}<{\centering}m{0.9cm}<{\centering}m{1.0cm}<{\centering}m{1.1cm}<{\centering}}
\toprule 
\bf Pruthi & Adv. Acc &ASR\% & \#Query & \#Adv\\
\toprule 
SST-2 & 59.2 & 36.0 & 326.9 & 111\\
Yelp & 86.4 & 11.5 & 1678.1 & 1036\\
AGNews & 84.5 & 11.1 & 792.0 & 239\\
SNLI & 23.2 & 74.4 & 103.4 & 1846\\
\bottomrule\hline
\end{tabular}
\begin{tabular}{m{1.3cm}<{\centering}|m{1.3cm}<{\centering}m{0.9cm}<{\centering}m{1.0cm}<{\centering}m{1.1cm}<{\centering}}
\toprule 
\bf TextBugger & Adv. Acc &ASR\% & \#Query & \#Adv\\
\toprule 
SST-2 & 28.9 & 68.7 & 49.3 & 221\\
Yelp & 16.3 & 83.3 & 350.1 & 738\\
AGNews & 20.2 & 79.2 & 123.4 & 1088 \\
SNLI & 4.5 & 95.0 & 41.9 & 2225\\
\bottomrule\hline
\bottomrule
\end{tabular}
\end{center}
\caption{Statistics about attacks. We report the adversarial accuracy (Adv. Acc), attack success rate (ASR\%), the number of queries (\#Query), and the number of adversarial examples examined.}
\label{appendix:attacks}
\end{table}

\begin{table}[t]
\renewcommand{\arraystretch}{0.85}
\begin{center}
\small
\begin{tabular}{m{1.3cm}<{\centering}|m{1.3cm}<{\centering}m{0.9cm}<{\centering}m{1.0cm}<{\centering}m{1.1cm}<{\centering}}
\toprule
\toprule 
\bf TF-FB & Adv. Acc &ASR\% & \#Query & \#Adv\\
\toprule 
SST-2 & 6.54 & 94.8 & 108.4 & 295\\
Yelp & 6.2 & 93.7 & 496.0 & 1027\\
AGNews & 22.0 & 77.4 & 365.7 & 1604\\
SNLI & 8.3 & 91.4 & 69.6 & 2068\\
\bottomrule\hline
\end{tabular}
\begin{tabular}{m{1.3cm}<{\centering}|m{1.3cm}<{\centering}m{0.9cm}<{\centering}m{1.0cm}<{\centering}m{1.1cm}<{\centering}}
\toprule 
\bf BAE-FB & Adv. Acc &ASR\% & \#Query & \#Adv\\
\toprule 
SST-2 & 45.3 & 52.2 & 64.3 & 164 \\
Yelp & 50.2 & 48.8 & 323.4 & 333\\
AGNews & 87.6 & 9.7 & 119.5 & 202\\
SNLI & 46.8 & 51.3 & 44.5 & 1347\\
\bottomrule\hline
\end{tabular}
\begin{tabular}{m{1.3cm}<{\centering}|m{1.3cm}<{\centering}m{0.9cm}<{\centering}m{1.0cm}<{\centering}m{1.1cm}<{\centering}}
\toprule 
\bf Pruthi-FB & Adv. Acc &ASR\% & \#Query & \#Adv\\
\toprule 
SST-2 & 68.9 & 27.3 & 326.4 & 90\\
Yelp & 89.4 & 9.1 & 1681.0 & 134\\
AGNews & 89.8 & 7.4 & 791.4 & 158\\
SNLI & 47.2 & 50.9 & 103.8 & 1323\\
\bottomrule\hline
\end{tabular}
\begin{tabular}{m{1.3cm}<{\centering}|m{1.3cm}<{\centering}m{0.9cm}<{\centering}m{1.0cm}<{\centering}m{1.1cm}<{\centering}}
\toprule 
\bf TB-FB & Adv. Acc &ASR\% & \#Query & \#Adv\\
\toprule 
SST-2 & 35.3 & 62.6 & 53.0 & 207\\
Yelp & 18.4 & 81.3 & 369.4 & 1025\\
AGNews & 53.1 & 45.3 & 191.1 & 948 \\
SNLI & 18.1 & 81.2 & 50.1 & 2093\\
\bottomrule\hline
\bottomrule
\end{tabular}
\end{center}
\caption{Statistics about FB attacks. We report the adversarial accuracy (Adv. Acc), attack success rate (ASR\%), the number of queries (\#Query), and the number of adversarial examples examined.}
\label{appendix:attackshp}
\end{table}

\section{DIST}
\label{Appendix:dist}
We propose the \textbf{DIST} baseline, which is a distance-based detector motivated by \citet{Ma2018CharacterizingAS}. We also find that the Local Intrinsic Dimension value proposed in \citet{Ma2018CharacterizingAS} does not work well when detecting NLP attacks. The DIST method leverages the whole training set as $\mathcal{D}_{aux}$. Then, it selects the K-nearest neighbors of the evaluated point from each class of $\mathcal{D}_{aux}$ and calculates the average distance between the neighbors and the evaluated point, denote as $d_1, d_2, \cdots, d_k$, where $k$ is the number of classes. Suppose the evaluated point has predicted class $i$. Then, it uses the difference between the distance of class $i$ and the minimum of the other classes to do detection. i.e., $d_i - min\left(d_k\right)$, where $k \neq i$. The intuition is that since adversarial examples are generated from the original class, they might still be closer to training data in the original class, which is $min\left(d_k\right), k \neq i$. 

\section{Implementation Details}
\label{Appendix:implement}
For DIST and ADDMU, we tune the hyperparameters with an attacked validation set. For datasets with an original train/validation/test split (SNLI), we simply attacked the examples in the validation set and select 100 of them to help the tuning. For datasets without an original split, like SST-2, Yelp, and AGNews, we randomly held out 100 examples from the training set and attack them to construct a set for hyperparameter tuning. For DIST, we select the number of the neighbors from \{100, 200, $\cdots$, 1000\}. For ADDMU, we select $N_m$ and $N_d$ from \{10, 20, 80, 100\}, and choose $p_m$ and $p_d$ from \{0.1, 0.2, 0.3, 0.4\}. In our preliminary experiment, we find that ensemble different $p_d$ values also help. So, we also consider ensemble different $p_d$ values in combinations \{(0.1, 0.2), (0.1, 0.2, 0.3, 0.4)\}. We also find that augment the model uncertainty estimation with some neighborhood data is helpful, so for the model uncertainty value, we actually average over 10 neighborhood data with 0.1 mask rate.

\begin{table}[t]
    \centering
    \begin{tabular}{c|c|c|c|c}
    \toprule 
      & TF & TF-FB & BAE & BAE-FB \cr
PPL & 75.8 & 77.1 & 41.9 & 40.9 \cr
FGWS & 86.2 & 87.1 & 83.7 & 81.4 \cr
RDE & 15.6 & 24.0 & 21.2 & 33.3 \cr
ADDMU & 93.7 & 89.3 & 92.2 & 87.6 \cr
\bottomrule
    \end{tabular}
    \caption{Detection results on a BiLSTM victim model. The values are F1 score on FPR=0.1. We see that ADDMU still achieves the best performance on these two attacks. Note also that RDE, the previous SOTA results on BERT and RoBERTa actually breaks when trying to detect BiLSTM adversarial examples.}
    \label{tab:lstm}
\end{table}

\begin{table}[t]
\renewcommand{\arraystretch}{0.85}
\begin{center}
\small
\begin{tabular}{R{1.6cm} p{1.20cm} p{0.70cm} p{1.20cm} p{0.80cm}}
% & \multicolumn{2}{c}{\bf{PTB-SD3.5}}
\toprule 
SST-2 TF &  \textbf{Clean \%} & \textbf{\#Aug}& \textbf{ASR} & \textbf{\#Query} \\
\toprule 
\multicolumn{1}{l}{BERT} &  95.8 & 0 & 88.66\% & 118.99 \\
+ All   & 95.6 & 11199 & 77.58\% & 140.74 \\
+ LDLM  & 95.6 & 2800& 82.52\% & 137.50 \\
+ HDLM  & 95.8 & 2800& 78.25\% & 142.26 \\
+ LDHM  & \textbf{95.8} & 2799& \textbf{75.30\%} & \textbf{145.79} \\
+ HDHM  & 95.3 & 2799 & 77.67\% & 142.42 \\
\bottomrule\hline
\end{tabular}
\end{center}
\caption{ADA performances for FB version of different types of augmented data. We find that adversarial examples with low DU and high MU are most useful for ADA.}
\label{appendix:FBADATest}
\end{table}

\begin{table}[t]
\renewcommand{\arraystretch}{0.85}
\begin{center}
\small
\begin{tabular}{p{1.6cm} p{1.20cm} p{0.70cm}}
% & \multicolumn{2}{c}{\bf{PTB-SD3.5}}
\toprule 
 &  Regular & FB \\
\toprule 
Regular &  87.2 & 90.2  \\
FB   & 82.3 & 77.1 \\
\bottomrule\hline
\end{tabular}
\end{center}
\caption{Attack success rate for four settings of augmentation. The columns are the augmented data. The rows are the attack types.}
\label{appendix:FBvsreg}
\end{table} 

\section{Selecting Useful data with Uncertainty values}
\label{Appendix:ADA}
In this section, we present the results of selecting useful data for ADA using DU and MU values for FB version of TF, shown in Table \ref{appendix:FBADATest}. Similar to the regular version, we find that the most useful data are still those with low data uncertainty and high model uncertainty. We achieve better ASR and the number of queries using only one quarter of data compared to the full augmentation. In Table \ref{appendix:FBvsreg}, we show the attack success rate of four settings. 1) Augment with FB examples to defend against regular attack; 2) Augment with FB examples to defend against FB attack;
3) Augment with regular examples to defend against regular attack;
4) Augment with regular examples to defend against FB attack. The finding is that augment with FB and regular adversarial examples most benefits its own attacks. This implies that FB attacks might already change the characteristics of regular attacks. We need to defend against them with different strategies.

\section{RoBERTa Results}
\label{Appendix:Roberta}
We conduct adversarial examples detection with RoBERTa-base. The setting is the same as BERT. Through hyperparameters search as described before, for ADDMU, we select $N_m=20$ and $N_d=100$, and choose $p_m=0.1$ and $p_d=0.1$, without augmentation for MU estimation and no ensemble of various $p_d$. Table \ref{Appendix:robertatable} presents the results for RoBERTa-base. ADDMU also outperform other methods with RoBERTa. We combine the ablation table together with the main table for RoBERTa.

\section{Ablation Study}
\label{appendix:ablation}
We present the full results for the ablation study of uncertainty aggregation in Table \ref{appendix:ablationtable}. We also show that our neighborhood construction process in data uncertainty can be used to enhance two baselines RDE and DIST.

\begin{table*}[!t]
\renewcommand{\arraystretch}{0.8}
\centering
\small
\begin{tabular}{m{1.58cm}|m{1.38cm}|m{.55cm}<{\centering}m{.55cm}<{\centering}m{.65cm}<{\centering}|m{.55cm}<{\centering}m{.55cm}<{\centering}m{.65cm}<{\centering}|m{.55cm}<{\centering}m{.55cm}<{\centering}m{.65cm}<{\centering}|m{.55cm}<{\centering}m{.55cm}<{\centering}m{.65cm}<{\centering}}
  \toprule
  &  & \multicolumn{3}{c}{\bf{SST-2}} & \multicolumn{3}{c}{\bf{AGNews}} & \multicolumn{3}{c}{\bf{Yelp}} &
  \multicolumn{3}{c}{\bf{SNLI}}
\\
\cmidrule{1-6}\cmidrule{7-10}\cmidrule{11-14}
   \bf{Attacks} & \bf{Methods} & \bf{TPR}  & \bf{F1} & \bf{AUC} & \bf{TPR}  & \bf{F1} & \bf{AUC} & \bf{TPR}  & \bf{F1} & \bf{AUC}  & \bf{TPR}  & \bf{F1} & \bf{AUC}\\
  \toprule
  \multirow{8}{4em}{{TF}} &
 RDE&62.9& 72.8& 86.5& 96.0& 93.2& 97.0& 72.0& 79.2& 89.6& 46.3& 59.3& 81.0 \cr 
 &RDE-aug&63.6& 73.3& 86.6& 97.4& 94.0& 97.4& 70.1& 77.9& 89.8& 41.0& 54.3& 79.9 \cr 
 &DIST&64.0& 73.4& 87.9& 94.5& 92.4& 95.9& 73.8& 80.3& 90.6& 37.2& 50.4& 74.5 \cr 
 &DIST-aug&60.2& 70.5& 86.5& 94.0& 92.0& 96.9& 75.7& 81.4& 90.8& 38.3& 51.5& 75.2 \cr 
 &MU&51.9& 64.2& 85.9& 82.0& 85.4& 94.5& 71.7& 79.0& 90.1& 65.1& 74.4& 89.1 \cr 
 &DU&60.6& 71.1& 87.8& 98.9& 94.6& 98.3& 76.3& 82.0& 90.6& 59.6& 70.3& 85.6 \cr 
 &ADDMU&\bf 67.1&\bf 75.8&\bf 88.8&\bf 99.2&\bf 94.9&\bf 98.6& \bf 78.7& \bf 83.5&\bf 91.6&\bf 68.9&\bf 77.0& \bf 89.7 \cr 

  \midrule
   \multirow{8}{4em}{{TF-FB}} &
 RDE&31.9& 45.0& 81.5& 71.9& 79.1& 92.5& 31.5& 44.6& 82.7& 43.1& 56.4& 79.6 \cr 
 &RDE-aug&36.6& 50.2& 80.4& 90.8& 90.5& 95.9& 61.5& 71.8& 87.8& 37.6& 51.0& 78.9 \cr 
 &DIST&20.7& 26.3& 81.6& 66.6& 75.4& 91.8& 54.8& 64.3& 86.2& 27.2& 39.6& 69.9 \cr 
 &DIST-aug&50.5& 62.4& 84.0& 81.9& 85.3& 94.5& 64.0& 73.5& 88.5& 29.6& 42.3& 71.0 \cr 
  &MU&54.4& 66.3& 85.4& 90.7& 90.4& 96.5& 70.7& 78.3& 89.1& \bf 60.2& \bf 70.8& 87.0 \cr 
 &DU&55.4& 67.1& 84.5& 97.2& 93.9& 97.5& 70.1& 77.9& 88.4& 31.6& 44.6& 78.2 \cr 
 &ADDMU&\bf 59.4& \bf70.6&\bf 87.3&\bf 97.5&\bf 94.0&\bf 97.8& \bf72.8&\bf 79.7&\bf 89.7&53.6& 65.8&\bf 87.5 \cr 

  \midrule
  \midrule
   \multirow{8}{4em}{{BAE}} &
 RDE&44.2& 57.3& 79.3& 96.4& 93.7& 96.3& 65.2& 74.5& 89.1& 41.7& 55.0& 76.8 \cr 
 &RDE-aug&\bf 49.3&\bf 61.9&82.4& 85.6& 87.8& 94.3& 61.7& 71.9& 88.5& 44.9& 57.9& 80.3 \cr 
 &DIST&44.9& 57.3& 78.9& 94.2& 91.9& 96.2& 68.0& 76.2& 89.4& 36.8& 49.7& 67.9 \cr 
 &DIST-aug&38.1& 50.7& 77.8& 86.3& 87.9& 94.7& 66.1& 74.8& 89.7& 38.3& 51.6& 69.8 \cr 
  &MU&41.7& 55.0& 78.8& 86.7& 88.3& 94.1& 64.6& 74.1& 88.6& 44.4& 57.5& 76.9 \cr 
 &DU&45.9& 58.9&\bf  83.3&\bf 97.5&\bf 94.3&\bf 98.1& 71.5& 78.5& 89.7& 44.7& 57.8& 80.5 \cr 
 &ADDMU&45.9& 58.9& 82.3& 96.4& 93.5& 97.3&\bf 72.5&\bf 79.5& \bf 90.1&\bf 48.2&\bf 61.0&\bf 81.0 \cr 

   \midrule
   \multirow{8}{4em}{{BAE-FB}} &
 RDE&19.5& 30.2& 72.5& 68.8& 77.0& 91.2& 66.4& 75.4& 88.1& 34.6& 47.9& 74.0 \cr 
 &RDE-aug&48.2& 61.0& 82.6& 63.4& 73.4& 91.1& 66.1& 75.1& 88.9& 40.8& 54.2& 79.4 \cr 
 &DIST&17.7& 26.1& 70.1& 64.9& 68.1& 91.4& 69.7& 77.3& 88.4& 29.5& 42.3& 62.9 \cr 
 &DIST-aug&28.7& 40.0& 72.4& 70.3& 76.3& 91.6& 71.5& 78.0& 89.8& 31.5& 44.0& 65.4 \cr 
  &MU&49.7& 62.3& 82.3& 83.7& 86.4& 94.0& 74.5& 80.8& 89.9& \bf 36.5&\bf 49.9&73.1 \cr 
 &DU&\bf 56.4&\bf 67.8& 84.4&\bf 84.7&\bf 87.0& 93.4& 74.5& 80.8& 90.2& 22.9& 34.5& 74.3 \cr 
 &ADDMU&51.4& 64.1&\bf 84.6& 83.7& 85.9& \bf 94.1&\bf 76.3& \bf81.9&\bf 90.6& 34.9& 48.4& \bf 76.0 \cr 

   \midrule
  \midrule
   \multirow{8}{4em}{{Pruthi}} &
 RDE&41.4& 55.1& 80.6& 77.4& 82.8& 92.4& 52.6& 64.8& 88.0& 34.6& 47.8& 76.5 \cr 
 &RDE-aug&40.5& 53.9& 78.9& 87.4& 88.7& 94.1& 64.7& 74.3& 88.0& 35.4& 48.7& 77.3 \cr 
 &DIST&55.0& 61.4& 82.9& 77.8& 82.0& 92.1& 66.7& 72.2& 88.2& 23.6& 35.2& 65.1 \cr 
 &DIST-aug&50.5& 61.4& 84.1& 81.2& 84.6& 94.1& 69.2& 75.6& 89.5& 26.4& 38.7& 67.4 \cr 
  &MU&48.6& 61.4& 85.3& 89.5& 89.9& 95.5& 77.5& 83.1& \ 90.7&\bf 61.8&\bf 72.0&\bf 86.8 \cr 
 &DU& 55.7&  66.8& 82.7& 95.8& 93.8& 97.3& 72.4& 79.6& 88.8& 26.6& 39.0& 74.4 \cr 
 &ADDMU&\bf 55.9&\bf 67.4&\bf 85.4& \bf 96.7&\bf 93.9&\bf 97.4&\bf 78.8& \bf 83.7&\bf 91.8& 55.7& 67.1& 86.0 \cr 

   \midrule
   \multirow{8}{4em}{{Pruthi-FB}} &
 RDE&20.0& 30.8& 72.6& 59.5& 70.4& 87.6& 34.3& 47.9& 85.2& 31.2& 44.2& 74.9 \cr 
 &RDE-aug&26.7& 39.0& 74.5& 67.7& 76.4& 91.8& 60.4& 71.1& 87.0& 31.0& 44.0& 76.0 \cr 
 &DIST&23.3& 26.5& 74.6& 55.1& 61.6& 87.2& 54.5& 55.2& 84.9& 21.6& 32.8& 63.3 \cr 
 &DIST-aug&25.6& 35.0& 76.0& 69.6& 76.3& 91.3& 59.7& 69.6& 87.5& 23.8& 35.4& 65.7 \cr 
  &MU&56.2& 67.7& 85.2& 80.3&84.8& 94.5& 67.9& 76.8& 91.7&\bf 60.7&\bf 71.1&\bf 85.5 \cr 
 &DU&56.2& 68.5& 83.1& 79.1& 83.9& 93.5& 67.2& 75.9& 86.4& 13.9& 22.4& 70.3 \cr
 &ADDMU&\bf 56.2&\bf 68.7&\bf 85.8&\bf 80.4&\bf 84.9&\bf 95.0&\bf 68.7& \bf 77.0&\bf 91.7& 44.9& 58.0& 82.5 \cr 
 
   \midrule
   \midrule
   \multirow{8}{4em}{{TB}} &
 RDE&72.4& 79.6& 89.6& 96.1& 93.3& 96.9& 66.2& 75.2& 89.2& 51.8& 64.1& 83.0 \cr 
 &RDE-aug&54.3& 66.1& 85.0& 95.6& 93.0& 96.9& 61.7& 71.9& 87.8& 45.9& 58.9& 80.9 \cr 
 &DIST&72.4& 78.6& 90.6& 95.6& 92.8& 96.2& 70.2& 77.9& 90.2& 50.7& 62.7& 82.6 \cr 
 &DIST-aug&72.9& 79.1& 89.7& 93.0& 91.6& 96.3& 70.5& 78.0& 90.5& 52.0& 64.2& 83.1 \cr 
  &MU&67.4& 76.0& 88.9& 79.8& 84.1& 94.5& 67.0& 75.7& 88.9& 60.2& 70.8& 88.6 \cr 
 &DU&\bf 77.8&\bf 82.9& 90.2& 98.4& 94.7& 98.0& 69.3& 77.3& 89.2& 66.9& 75.4& 88.9 \cr 
 &ADDMU&73.3& 80.0&\bf 90.9&\bf 99.0&\bf 94.8&\bf 98.4&\bf 70.8&\bf 78.3&\bf 91.0&\bf 69.0&\bf 77.1&\bf 90.6 \cr 

    \midrule
   \multirow{8}{4em}{{TB-FB}} &
 RDE&29.5& 42.5& 82.1& 68.9& 77.1& 91.7& 63.9& 73.5& 88.4& 47.8& 60.6& 82.2 \cr 
 &RDE-aug&42.0& 55.4& 80.2& 86.6& 88.2& 94.7& 59.6& 70.3& 87.5& 40.7& 54.0& 80.1 \cr 
 &DIST&34.3& 44.0& 82.6& 63.4& 72.9& 91.5& 69.8& 77.6& 89.3& 40.8& 53.9& 79.0 \cr 
 &DIST-aug&49.8& 59.0& 84.6& 80.4& 84.3& 93.6& 71.8& 78.9& 90.4& 43.9& 57.0& 79.8 \cr 
  &MU&55.9& 67.4& 85.8& 91.8& 91.0& 96.1& 72.2& 79.4& 89.6& \bf 57.7&\bf 68.8& 87.0 \cr 
 &DU&\bf 58.1&\bf 69.2& 85.0& 94.1& 92.2& 96.5& 72.7& 79.6& 89.2& 40.9& 54.2& 81.5 \cr 
 &ADDMU&50.5& 62.9&\bf 86.1&\bf 94.2&\bf 92.6&\bf 96.9&\bf 74.8&\bf 81.0&\bf 90.8& 51.1& 63.6&\bf 87.0 \cr

  \bottomrule\hline

\end{tabular}
\caption{Ablation of detection performance of regular and FB adversarial examples (*-FB) against BERT on SST-2, AGNews, Yelp, and SNLI. We compare ADDMU with soley DU, solely MU, and two enhanced baselines RDE-aug and DIST-aug. The best performance is bolded. Results are averaged over three runs with different random seeds.}
\label{appendix:ablationtable}
\end{table*}

\begin{table*}[!t]
\renewcommand{\arraystretch}{0.8}
\centering
\small
\begin{tabular}{m{1.58cm}|m{1.58cm}|m{.75cm}<{\centering}m{.75cm}<{\centering}m{.75cm}<{\centering}|m{.75cm}<{\centering}m{.75cm}<{\centering}m{.75cm}<{\centering}}
  \toprule
  &  & \multicolumn{3}{c}{\bf{SST-2}} & \multicolumn{3}{c}{\bf{AGNews}} 
\\
\cmidrule{1-8}
   \bf{Attacks} & \bf{Methods} & \bf{TPR}  & \bf{F1} & \bf{AUC} & \bf{TPR}  & \bf{F1} & \bf{AUC} \\
  \toprule
  \multirow{10}{4em}{{TF}} &
 PPL&34.0& 47.2& 73.7& 78.2& 83.1& 92.0 \cr 
  &MSP&71.0& 78.5& 89.8& 93.5& 91.9& 97.2 \cr 
 &RDE&73.9& 80.4& 89.8& 90.6& 90.4& 95.5 \cr 
 &RDE-aug&61.3& 71.6& 87.1& 61.7& 71.9& 87.9 \cr 
 &DIST&70.3& 77.9& 90.2& 94.6& 92.5& 96.5 \cr 
 &DIST-aug&72.7& 78.8& 90.1& 83.8& 86.4& 94.6 \cr 

 &MU&78.0& 82.9& 91.1& 98.7& 94.6& 97.6 \cr 
 &DU&70.1& 79.2& 89.5& 95.9& 93.2& 97.6 \cr 
 &ADDMU&\bf 78.4&\bf 83.9&\bf 91.3&\bf 98.8&\bf 94.9&\bf 98.3 \cr 
  \midrule
   \multirow{10}{4em}{{TF-FB}} &
 PPL&43.8& 57.0& 79.6& 84.4& 86.9& 94.1 \cr 
 &MSP&55.3& 66.9& 85.0& 30.5& 43.5& 87.6 \cr 
 &RDE&40.5& 53.8& 84.6& 57.0& 68.3& 88.7 \cr 
 &RDE-aug&46.7& 59.7& 82.4& 48.1& 60.9& 81.9 \cr 
 &DIST&48.7& 58.2& 85.1& 47.0& 59.7& 89.8 \cr 
 &DIST-aug&48.7& 60.8& 85.8& 67.4& 75.4& 90.9 \cr 

 &MU&\bf55.9& \bf66.9&\bf 89.2& 77.4& 82.6& 93.5 \cr 
 &DU&55.1& 64.2& 84.5& 88.4&\bf 89.6& 95.7 \cr 
  &ADDMU&54.6& 66.8& 88.5&\bf 88.6& 89.2&\bf 95.8 \cr 
  \midrule
  \midrule
   \multirow{10}{4em}{{BAE}} &
 PPL&17.2& 27.1& 64.0& 38.1& 51.5& 74.0 \cr 
  &MSP&48.1& 60.9& 78.6& 93.4& 91.9& 97.2 \cr 
 &RDE&53.8& 65.7& 80.3& 77.2& 82.5& 93.2 \cr 
 &RDE-aug&53.5& 65.5& 84.4& 52.9& 64.9& 82.6 \cr 
 &DIST&48.1& 60.3& 79.7& 88.3& 88.9& 95.0 \cr 
 &DIST-aug&48.5& 61.2& 80.9& 72.4& 79.0& 91.3 \cr 

 &MU&55.7& 66.9& 81.8& 93.7& 92.0& 95.9 \cr 
 &DU&52.4& 64.0& 84.6& 92.2& 91.2& 96.2 \cr 
  &ADDMU&\bf55.8&\bf 67.0&\bf 84.9&\bf 97.6&\bf94.1&\bf 97.9 \cr 
   \midrule
   \multirow{10}{4em}{{BAE-FB}} &
 PPL&27.0& 39.6& 68.7& 34.5& 47.8& 73.6 \cr 
  &MSP&31.4& 44.6& 69.8& 77.0& 82.4& 89.8 \cr 
 &RDE&25.8& 38.1& 72.5& 57.0& 68.3& 89.4 \cr 
 &RDE-aug&40.3& 53.8& 77.9& 61.6& 71.9& 89.7 \cr 
 &DIST&25.8& 31.2& 71.2& 36.0& 47.2& 89.9 \cr 
 &DIST-aug&30.8& 42.7& 73.8& 62.0& 70.2& 89.5 \cr 
 
 &MU&43.4& 56.2& 76.5& 92.0& 91.3& 95.0 \cr 
 &DU&37.7& 51.3& 78.1& 79.5& 86.9& 93.3 \cr 
  &ADDMU&\bf44.4&\bf 57.1&\bf 78.3&\bf 92.0&\bf 91.9&\bf 95.7 \cr
   \midrule
  \midrule
   \multirow{10}{4em}{{Pruthi}} &
 PPL&34.0& 47.6& 74.4& 31.4& 44.4& 73.9 \cr 
  &MSP&62.0& 72.1& 83.1& 70.2& 78.0& 93.0 \cr 
 &RDE&57.0& 68.3& 83.3& 61.6& 71.9& 89.7 \cr 
 &RDE-aug&52.0& 64.2& 84.4& 40.8& 54.2& 81.1 \cr 
 &DIST&63.0& 70.6& 83.1& 65.5& 74.6& 92.5 \cr 
 &DIST-aug&52.0& 64.6& 84.3& 72.2& 77.5& 91.1 \cr 

 &MU&73.0& 77.1& \bf89.3&\bf 94.5&\bf 92.5& 97.5\cr 
 &DU&58.0& 68.3& 84.9& 85.1& 87.3& 95.5 \cr
  &ADDMU&\bf77.0& \bf82.4& 88.0& 92.5& 91.5& \bf97.6 \cr 
   \midrule
   \multirow{10}{4em}{{Pruthi-FB}} &
 PPL&23.4& 35.3& 71.0& 27.9& 40.6& 71.5 \cr 
  &MSP&40.6& 54.2& 76.3& 12.5& 20.5& 83.5 \cr 
 &RDE&51.6& 64.1& 78.4& 35.3& 48.7& 82.1 \cr 
 &RDE-aug&37.5& 51.1& 75.7& 26.5& 39.1& 74.6 \cr 
 &DIST&43.8& 22.8& 77.3& 25.5& 34.4& 83.4 \cr 
 &DIST-aug&40.6& 47.8& 79.6& 56.6& 67.0& 86.1 \cr 

 &MU&64.1& 62.9& 84.9& 72.8& 78.4& 92.8 \cr 
 &DU&39.1& 51.1& 81.1& 61.8& 71.6& 90.9 \cr 
  &ADDMU&\bf 64.1&\bf 74.5&\bf 89.1&\bf 75.7&\bf 82.1&\bf 93.4 \cr 
   \midrule
   \midrule
   \multirow{10}{4em}{{TB}} &
 PPL&45.5& 58.6& 81.2& 76.7& 82.2& 91.2 \cr 
  &MSP&74.7& 81.1& 91.8& 91.4& 90.8& 96.8 \cr 
 &RDE&76.8& 82.4& 92.0& 86.0& 87.8& 84.5 \cr 
 &RDE-aug&60.6& 71.2& 88.4& 57.4& 68.6& 85.6 \cr 
 &DIST&76.3& 80.7& 91.5& 93.4& 91.7& 96.0 \cr 
 &DIST-aug&77.3& 80.1& 91.8& 80.1& 84.0& 93.9 \cr 

 &MU&78.3& 82.3& 92.4& 98.3& 94.4& 97.3 \cr 
 &DU&74.2& 80.4& 90.7& 94.0& 92.1& 97.0 \cr 
 &ADDMU&\bf78.8& \bf82.9&\bf 92.4& \bf98.3&\bf 94.5&\bf 97.9 \cr 
    \midrule
   \multirow{10}{4em}{{TB-FB}} &
 PPL&42.7& 55.9& 81.6& 78.9& 83.7& 92.6 \cr 
  &MSP&57.5& 69.4& 86.3& 29.8& 42.7& 87.6 \cr 
 &RDE&52.0& 64.3& 86.2& 47.7& 60.6& 87.3 \cr 
 &RDE-aug&45.6& 58.6& 81.8& 43.5& 56.8& 78.3 \cr 
 &DIST&48.5& 56.5& 86.1& 45.2& 58.0& 89.4 \cr 
 &DIST-aug&48.0& 59.7& 85.5& 60.0& 70.6& 89.4 \cr 

 &MU&58.5& 70.0&\bf 89.7& 84.2& 86.8& 95.0 \cr 
 &DU&53.9& 64.2& 84.6& 81.9& 84.5& 92.8 \cr 
   &ADDMU&\bf64.9& \bf74.2& 87.7&\bf 85.5&\bf 88.3&\bf 96.8 \cr 
  \bottomrule\hline

\end{tabular}
\caption{Detection performance of regular and FB adversarial examples (*-FB) against RoBERTa on SST-2, AGNews. Our proposed ADDMU outperforms other methods. The best performance is bolded. Results are averaged over three runs with different random seeds.}
\label{Appendix:robertatable}
\end{table*}

\section{Preliminary Results on BiLSTM}
\label{Appendix:LSTM}
We experiment with a one-layer BiLSTM model with hidden dimension 150 and dropout 0.3. The model achieves 89.3 clean accuracy on SST-2. In our preliminary experiments, we test on detecting TextFooler and BAE attacks and their corresponding FB attacked examples. We compare our ADDMU detector with three baselines PPL, FGWS, and RDE. Results are shown on Table \ref{tab:lstm}. We show that ADDMU still achieves the best performance, while the previous SOTA on detecting BERT and RoBERTa adversarial examples, RDE, is corrupted when detecting BiLSTM adversarial examples.

\section{Related work in CV}
\label{Appendix:relatedwork}

\citet{Feinman2017DetectingAS} train a binary classifier using density estimation and Bayesian uncertainty estimation as features for detection. \citet{Li2021DetectingAE} replace DNNs with Bayesian Neural Networks, which enhance the distribution dispersion between natural and adversarial examples and benefit AED. \citet{Roth2019TheOA} use \emph{logodds} on perturbed examples as statistics to conduct detection.
Further, \citet{Athalye2018ObfuscatedGG} have similar observations with us concerning image attacks. They find that the distance-based feature, \emph{local intrinsic dimension} proposed in \citet{Ma2018CharacterizingAS} for AED fails when encounters FB adversarial examples.
\end{document}